%% file: main.tex
\documentclass{article}

\usepackage[final]{corl_2020} 

\usepackage{bm}
\usepackage{graphicx}
\usepackage{comment}
\usepackage{amsmath,amssymb} 
\usepackage{color}
\usepackage{subfig}
\usepackage{mathtools}
\usepackage{xfrac}
\usepackage{booktabs}
\usepackage{multirow}
\usepackage{nicefrac}
\usepackage{appendix}

\DeclareMathOperator*{\argmin}{arg\,min}

\newcommand{\rares}[1]{\textcolor{black}{#1}}

\title{Self-Supervised 3D Keypoint Learning \\ for Ego-Motion Estimation}

\author{\textbf{Jiexiong Tang$^{1, 2,*}$} \quad
        \textbf{Rareș Ambruș$^{1, *}$} \quad
        \textbf{Vitor Guizilini$^{1}$}\quad
        \textbf{Sudeep Pillai$^{1}$} \\
        \textbf{Hanme Kim$^{1}$}\quad
        \textbf{Patric Jensfelt$^{2}$}\quad
        \textbf{Adrien Gaidon$^1$} \quad
\and
$^1$ Toyota Research Institute\quad
$^2$ KTH Royal Institute of Technology\\
{\tt\small $^1$\{firstname.lastname\}@tri.global}\quad
{\tt\small $^2$firstname@kth.se}\quad
}

\begin{document}
\maketitle
\vspace*{-4mm}
\begin{abstract}
\input{latex/00abstract}
\end{abstract}
\keywords{Self-supervised-learning, Keypoints, Monocular, Visual odometry}

\let\thefootnote\relax\footnotetext{$^*$Equal contribution. This work was part of an internship stay at TRI.}
\let\thefootnote\relax\footnotetext{$^\dagger$Video: \href{https://youtu.be/bWqGU9zoH9I}{https://youtu.be/bWqGU9zoH9I}} 

\input{latex/01introduction.tex}	
\input{latex/02related_work.tex}	
\input{latex/030procedure.tex}

\input{latex/040experiments.tex}

\input{latex/05results.tex}

\input{latex/06conclusion.tex}

\appendix

\input{latex/070suppmat.tex}

\bibliography{main}
\clearpage
\end{document}

%% file: latex/00abstract.tex
Detecting and matching robust viewpoint-invariant keypoints is critical for visual SLAM and Structure-from-Motion. State-of-the-art learning-based methods generate training samples via homography adaptation to create 2D synthetic views with known keypoint matches from a single image.
This approach, however, does not generalize to non-planar 3D scenes with illumination variations commonly seen in real-world videos.
In this work, we propose self-supervised learning of depth-aware keypoints directly from unlabeled videos. We jointly learn keypoint and depth estimation networks by combining appearance and geometric matching via a differentiable structure-from-motion module based on Procrustean residual pose correction.
We describe how our self-supervised keypoints can be integrated into state-of-the-art visual odometry frameworks for robust and accurate ego-motion estimation of autonomous vehicles in real-world conditions.$^\dagger$


%% file: latex/01introduction.tex
\vspace{-1mm}
\section{Introduction}
\label{sec:intro}	
\vspace{-1mm}

\begin{figure}[t]
    \centering
     \includegraphics[width=0.95\columnwidth]{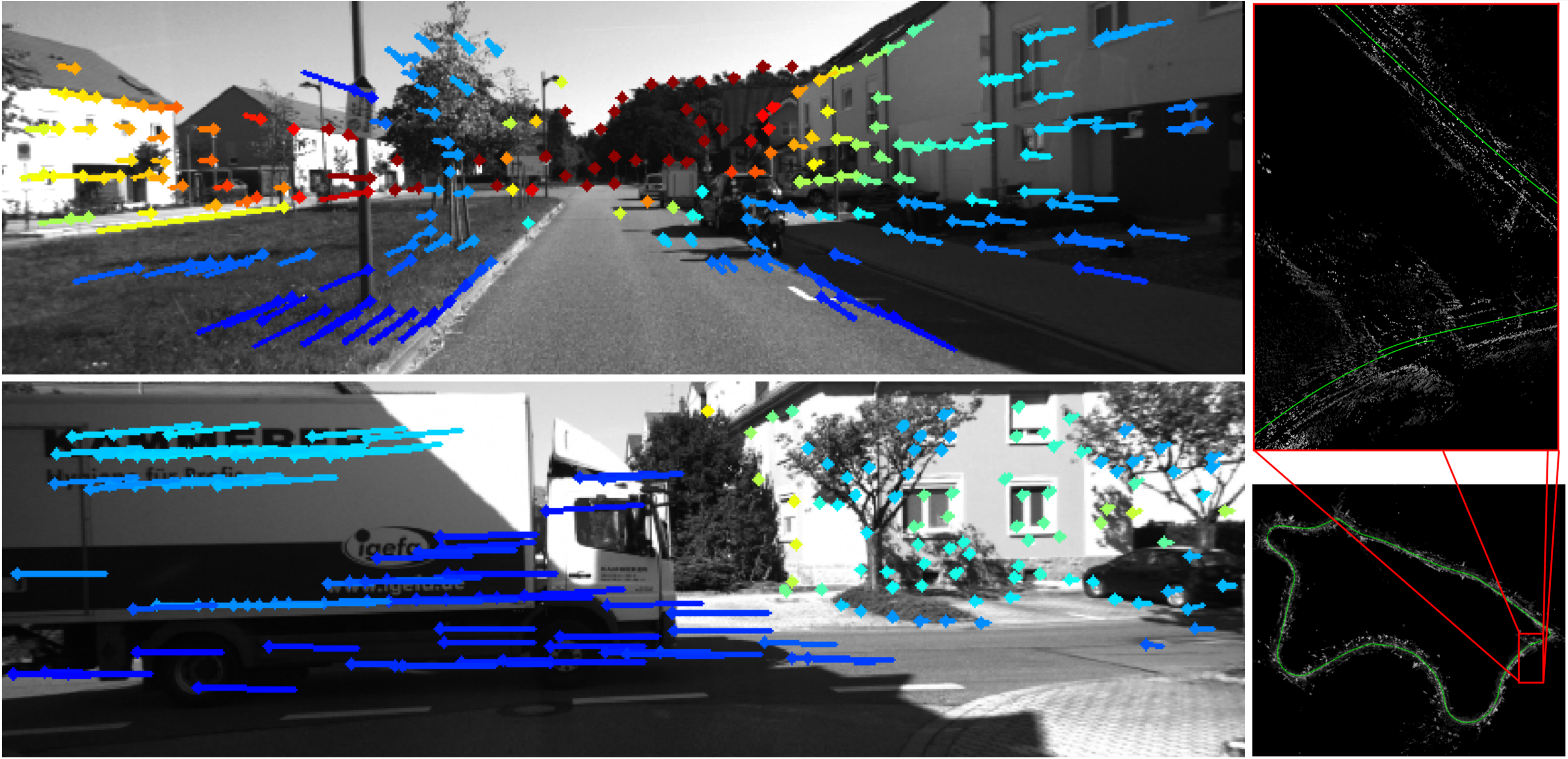}
    \caption{\textbf{Self-Supervised 3D Keypoints for Robust Visual-Odometry.}~\textbf{Left:}~The illustration shows the proposed self-supervised 3D keypoints learned \textit{purely} from unlabeled monocular videos together with matched sparse 3D scene flow. Our method can effectively handle dynamic objects via outlier rejection thanks to structured pose estimation with our 3D keypoints. 
    \textbf{Right:}~The proposed 3D keypoint estimator is integrated into a state-of-the-art visual tracking framework for accurate, scale-aware, long-range monocular visual odometry.}
    \label{fig:intro-fig}
\vspace{-3.5mm}
\end{figure}

Detecting interest points in images and matching them across views is a fundamental capability of many robotic systems. Tasks such as Structure-from-Motion (SfM)~\citep{agarwal2010bundle}, Visual Odometry (VO), and visual Simultaneous Localization and Mapping (SLAM)~\citep{cadena2016past} require salient keypoints to be detected and re-identified in diverse settings with strong invariance to lighting, viewpoint changes, and scale. Until recently, these tasks have relied on hand-engineered keypoint features~\citep{lowe1999object,rublee2011orb} with limited performance~\citep{sarlin2019coarse}. Deep learning has recently revolutionized many computer vision applications in the supervised setting~\citep{he2016deep,alp2018densepose,kirillov2019panoptic}. However, these methods rely on strong supervision in the form of ground-truth labels that are often expensive to acquire. Moreover, supervising interest point detection is challenging, as a human annotator cannot trivially identify salient regions in images that would allow their re-identification in diverse scenarios. 
Inspired by recent approaches to keypoint learning~\citep{detone2018superpoint,christiansen2019unsuperpoint,revaud2019r2d2,bhowmik2020reinforced}, we propose an approach that exploits the temporal context in videos to learn an accurate and repeatable monocular keypoint detector, descriptor, and 2D-3D keypoint lifting function in a fully \textit{self-supervised} manner. We focus on the task of \textit{ego-motion estimation} and show that by combining geometry and appearance in a joint optimization framework we can identify depth-aware monocular keypoints that are particularly well suited for pose estimation. In addition, we demonstrate that our 3D keypoint estimator can be effectively integrated into a state-of-the-art visual tracking framework for accurate, long-range monocular visual odometry (see Figure~\ref{fig:intro-fig}).

Our \textbf{main contribution} is a fully self-supervised framework for the learning of depth-aware keypoint detection and description purely from unlabeled videos. Our novel formulation allows us to simultaneously learn keypoint detection, matching, and 3D lifting for robust visual ego-motion. \rares{Our \textbf{second contribution} lies in our principled use of the Orthogonal Procrustes algorithm to differentially regress an SE3 pose that is tightly coupled with the joint estimation of depth and keypoints from videos}. We show that by enforcing strong regularization in the form of sparse multi-view geometric constraints, the keypoint and depth networks strongly benefit from joint optimization in an end-to-end framework. Finally, in our \textbf{third contribution}, we show results comparable to state-of-the-art stereo long-term tracking by integrating our self-supervised, monocular and depth-aware keypoints into existing visual tracking frameworks such as Direct Sparse Odometry (DSO)~\citep{engel2017direct}. 

%% file: latex/02related_work.tex
\vspace{-2mm}
\section{Related Work}
\label{sec:related}
\vspace{-2mm}

\textbf{Learning-based Methods for Keypoint Estimation.} Over the past two decades, handcrafted image features such as SIFT~\citep{lowe1999object} and ORB~\citep{rublee2011orb} have been the key enabler of visual SLAM~\citep{cadena2016past} and SfM applications~\citep{agarwal2010bundle}. More recently, learning-based keypoint detectors and descriptors have advanced the state-of-the-art performance on challenging benchmarks. While generating ground truth data is a tedious and expensive process,~\citet{detone2018superpoint} has shown that synthetic data can be used for supervising keypoints which can be transferred to real world scenes. Alternatively, SfM~\citep{yi2016lift} or two-view consistency~\citep{ono2018lf,suwajanakorn2018discovery} can been used for keypoint learning without any additional labels. Single images have also been successfully used to train generalizable keypoints, with explicit descriptor losses by~\citet{christiansen2019unsuperpoint} or without by~\citet{tang2019neural} . Additionally, descriptor discriminativeness can be learned, thus identifying high confidence matching regions~\citep{revaud2019r2d2}. Keypoints can also be learned in an end-to-end optimization for downstream tasks such as localization~\citep{Sarlin:etal:CVPR2019}, relative pose estimation~\citep{bhowmik2020reinforced,sarlin2020superglue} and 3D matching~\citep{bai2020d3feat}. Our work extends previous work~\citep{christiansen2019unsuperpoint,tang2019neural} by combining self-supervised keypoint learning with depth estimation in monocular videos and by using the temporal consistency and scene geometry to regress a robust and repeatable keypoint estimator.

\textbf{Learning-based Methods for Visual Odometry.} While many learning-based methods exist, we will focus our discussion on self-supervised methods. Recently,~\citet{godard2017unsupervised}  used stereo imagery to derive a photometric loss as proxy supervision to self-supervise a monocular depth network. This was extended to the generalized multi-view case by~\citet{zhou2017unsupervised}, leveraging SfM constraints to simultaneously learn depth and camera ego-motion from monocular image sequences. A number of end-to-end self-supervised methods have been proposed, using super-resolution~\citep{pillai2018superdepth}, learned~\citep{casser2018depth} or self-discovered object detectors~\citep{bian2019unsupervised}, two-stream rgb and depth networks~\citep{ambrus2019two}, optical flow~\citep{luo2019every,zhan2019visual,zhao2020towards}, modeling camera intrinsics~\citep{gordon2019depth}, using adversarial networks~\citep{feng2019sganvo}, etc. Additionally, a number of hybrid methods have been proposed that combine self-supervised learning of the depth network with traditional methods, for example using a robust estimator to compute the fundamental matrix and eliminate outliers~\citep{zhan2019visual,zhao2020towards} or by integrating learning~\citep{yang2018deep,yang2020d3vo}  with a state-of-the-art tracking method~\citep{engel2017direct}. Similar to~\citet{zhan2019visual,zhao2020towards}, we use traditional methods to take advantage of the model-based PnP solution and the inliers established to outfit a differentiable pose estimation module within the self-supervised 3D keypoint learning framework. Differentiable Procrustes for point cloud alignment has been used previously by~\citet{wang2019deep,choy2020deep}, with a focus on global registration as a supervised learning problem. \rares{In~\citep{murthy2019gradslam} RGB-D data is registered through differentiable nonlinear least squares, allowing gradient flow and enabling end-to-end differentiable SLAM.} In this paper we instead focus on self-supervised sparse keypoint detection and optimization for the task of ego-motion estimation, and show that our appearance-based depth-aware keypoints achieve superior results compared to other learning-based methods. 

%% file: latex/030procedure.tex
\vspace{-3mm}
\section{Self-Supervised Depth-Aware Keypoint Learning}
\label{sec:ss_3d_keypoint}	
\vspace{-1mm}

\input{latex/031procedure_notation.tex}

\input{latex/033procedure_pnp_pose.tex}

\input{latex/032procedure_learning.tex}

%% file: latex/031procedure_notation.tex
\begin{figure}[!t]
\centering
\includegraphics[width=0.85\textwidth]{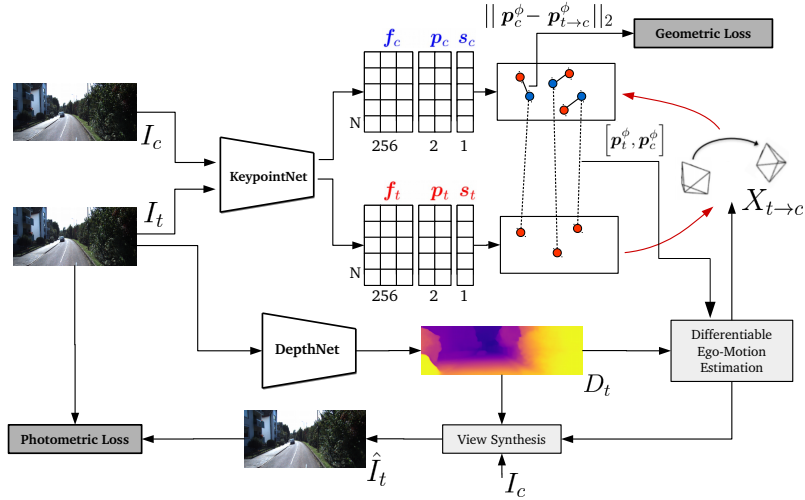}
\vspace{-22mm}
\caption{\textbf{Monocular SfM-based 3D Keypoint Learning}. We jointly optimize depth and keypoint networks thanks to differentiable view synthesis and geometric pose estimation modules. The whole system is self-supervised by strong geometric constraints on ego-motion thanks to our 3D keypoints.}
\label{fig:training_pipeline}
\vspace{-4mm}
\end{figure}

Inspired by the concept of leveraging known geometric transformations to self-supervise and boost keypoint learning~(\citet{detone2018superpoint}), we propose a novel self-supervised method that relies on epipolar constraints in two-view camera geometry for robust 3D keypoint learning. Crucially, we generalize previous work by~\citet{tang2019neural,christiansen2019unsuperpoint} and self-supervise 3D keypoint learning to leverage the structured geometry of scenes in unlabeled monocular videos, without any need for supervision in the form of ground-truth or pseudo-ground-truth labels. Thanks to learning the 2D-to-3D keypoint lifting function from monocular videos, we can accurately estimate the ego-motion between temporally adjacent images (see Figure~\ref{fig:training_pipeline} for an overview of the proposed pipeline).


\vspace{-2mm}
\subsection{Notation}
\label{subsec:notation}
\vspace{-2mm}

Our method works on sequences of monocular images, and we refer to $I_t$ as the target image and $I_C$ as the set of temporally adjacent context images, with $I_c \in I_C$.
We jointly learn two networks. First the \textbf{KeypointNet}~ $f_{k}: I \to  \left(\bm{p}, \bm{f}, \bm{s}\right)$ regresses $N$ \textit{image keypoints} consisting of positions $\bm{p} \in \mathbb{R}^{2\times N}$, descriptors $\bm{f} \in \mathbb{R}^{256\times N}$ and scores $\bm{s} \in \mathbb{R}^{N}$. Second, the \textbf{DepthNet}~ $f_D: I \to D$ predicts the \textit{scale-ambiguous} dense depth map $D$. We use $\bm{d} = D(\bm{p}) \in \mathbb{R}^{N}$ to denote the values of $D$ associated with the keypoint positions $\bm{p}=[\bm{u},\bm{v}]^{T}$. The \textbf{ego-motion estimator}~ $f_{x}(I_c,I_t)=X_{t \to c} = \begin{psmallmatrix}R & t\\ 0 & 1\end{psmallmatrix} \in \mathbb{SE}(3)$ predicts the relative 6-DoF rigid-body transformation between the target and context image. Throughout this manuscript, we use $\bm{p}$ to refer to sets of keypoints, while $p$ is used to refer to a single keypoint. Given two sets of detected keypoints, we define $\phi\left(\bm{f}_t,\bm{f}_c\right)$ as the \textbf{keypoint matching} function that computes keypoint correspondences via reciprocal matching in descriptor space:
\begin{equation} \label{eq:descriptor_matching}
\begin{gathered}
\phi\left(\bm{f}_t,\bm{f}_c\right) = \{\left(f_i,f_j\right) | \argmin_{j'}{\|f^2_i - f^2_{j'} \|_2} = j \land \argmin_{i'}{\|f^2_{i'} - f^2_j \|_2} = i \}
\end{gathered}
\end{equation}

with $f_i \in \bm{f}_t$ and $f_j \in \bm{f}_c$.
$\phi$ defines an association between the keypoint positions in the two images, which we denote by $\left[\bm{p}^{\phi}_t, \bm{p}^{\phi}_{c}\right]$. We use the correspondences $\left[\bm{p}^{\phi}_t, \bm{p}^{\phi}_c\right]$ and their corresponding predicted depths $\left[\bm{d}^{\phi}_t, \bm{d}^{\phi}_c\right]$ to estimate the rigid-body transformation $X_{t \to c}$ as described in detail in Section~\ref{subsec:pnp_pose}. Through $X_{t \to c}$, we define a joint optimization framework described in Section~\ref{subsec:learning}.

%% file: latex/033procedure_pnp_pose.tex
\subsection{Differentiable Pose Estimation from Depth-Aware Keypoints}
\label{subsec:pnp_pose}

\textbf{Pose Estimation via Perspective-n-Point.}
Using the estimated dense depth $D_t$ of the target image, we can compute the set $\bm{P}^{\phi}_t$ of 3D \emph{lifted} keypoints through the operation: $\bm{P}^{\phi}_t = \pi^{-1} (\bm{p}^{\phi}_t, \bm{d}^{\phi}_t)$ where $\bm{P}_t = [\bm{X}, \bm{Y}, \bm{Z}]^T \in \mathbb{R}^{3\times N}$ and $\pi(\cdot)$ is the standard pinhole camera projection model.
Using the keypoint correspondences $\left[\bm{p}^{\phi}_t, \bm{p}^{\phi}_c\right]$, we then have a 3D-2D correspondence set and can use the PnP algorithm~\cite{lepetit2009epnp} to compute the initial relative pose transformation $X^0_{t \to c}  = \begin{psmallmatrix}R_0 & t_0\\ 0 & 1\end{psmallmatrix}$ to geometrically match the keypoints in the target image to those in the context image. Specifically, we minimize:
\begin{equation} \label{eq:pnp}
E_{\psi}(X^{0}_{t \to c}) = \left\lVert \bm{p}^{\phi}_c - \pi\left( X^0_{t \to c} \cdot \bm{P}^{\phi}_t  \right)   \right\rVert_2~.
\end{equation}

The estimated relative pose $X^0_{t \to c}$ is obtained by minimizing the residual error in Equation~(\ref{eq:pnp}) using the Gauss-Newton (GN) method with RANSAC to ensure robustness to outliers (see Appendix~\ref{appendix:sec:pose_pnp} for details regarding this operation). This step allows us to compute the pose robustly, and yields an inlier set of correspondences $\left[\bm{p}^{\psi}_t, \bm{p}^{\psi}_c\right]$. However, this procedure is not differentiable with respect to the keypoint set used to estimate it. 

\textbf{Residual Pose Correction via Orthogonal Procrustes.} To alleviate the limitation of traditional PnP and allow end-to-end learning, we show how the initial pose estimate can be used to derive a 3D loss based on 3D-3D correspondences. We note that~\citet{sheffer2020pnp} have recently proposed a differentiable version of the traditional PnP algorithm which combines deep learning with a model-based fine-tuning step. While this method has shown good results when trained with noisy correspondences, we follow instead the monocular direct method of~\citet{engel2017direct} that performs frame-to-keyframe tracking. Specifically, we lift the context keypoints $\bm{p}^{\psi}_c$ to 3D using the re-projected depth of the source keypoints via the initial pose estimate $X^{0}_{t \to c}$. This allows us to form a 3D residual that can be used to recover the pose in closed-form (for convenience we omit the $\psi$ superscript below):
\begin{equation} \label{eq:3d_icp_1}
\begin{gathered}
E_{OP}(X_{t \to c}) = {\| \bm{P}_c - X_{t \to c} \cdot \bm{P}_t \|}_2~, \\
\text{where}~\mathbf{P}_t=\pi^{-1}(\bm{p}_t, D_t(\bm{p}_t))~,~\bm{P}_c=\pi^{-1}(\bm{p}_c, \bm{d}_c)~,~\text{and}~\bm{d}_c = \left[ X^0_{t \to c} \cdot \bm{P}_t \right]_{z}~. \\
\end{gathered}
\end{equation}
\noindent The 3D residual above can be effectively minimized by estimating the rotation and translation separately using a closed-form solution on the established inlier set.
We first estimate the rotation by subtracting the means of the points and minimizing Eq.~\ref{eq:orthogonal_residual} by solving an SVD in closed-form (otherwise known as the Orthogonal Procrustes problem~\cite{zhang2000flexible}):
\begin{equation} \label{eq:orthogonal_residual}
E(R) = \| \bm{P}^*_c - R \cdot \bm{P}^*_t \|_2~, \quad
\text{where} \enskip ~\bm{P}^*_i = \bm{P}_i - \bm{\overline{P}}_i~,
\end{equation}
\vspace{-4mm}
\begin{equation} \label{eq:3d_icp_2}
U\Sigma V = \mathtt{SVD}\left(\sum {(\bm{P}^*_c)}^T (\bm{P}^*_t)\right)~, \enskip \text{where} \quad R = V U^T~.
\end{equation}
Once the rotation $R$ is computed, the translation $t$ can be directly recovered by minimizing $t = \bm{P}^*_c - R \bm{\cdot} \bm{P}^*_t~$. Thus, the gradients for the pose rotation and translation can be effectively propagated with respect to the lifted 3D keypoint locations, making the overall pose estimation fully-differentiable. The differentiable pose estimated using the 2D keypoints from the source image and 3D keypoints from the target image tightly couples keypoint and depth estimation, thereby allowing both predictions to be further optimized using the overall keypoint learning objective described next. 

%% file: latex/032procedure_learning.tex
\subsection{Joint Self-Supervised Learning of Depth and Keypoints}
\label{subsec:learning}

We self-supervise the joint end-to-end learning of the keypoint and depth networks using a combination of a geometric keypoint loss $\mathcal{L}_{kpn}$, based on keypoint reprojection error, and a dense photometric loss $\mathcal{L}_{depth}$, based on the warped projection of $D_t$ in $I_c$: $\mathcal{L} = \mathcal{L}_{depth} + \alpha\mathcal{L}_{kpn}$.

\subsubsection{Keypoint Loss}

The total keypoint loss is composed of three terms: $\mathcal{L}_{kpn} = \mathcal{L}_{geom} + \beta_1\mathcal{L}_{desc} + \beta_2\mathcal{L}_{score}$.

\textbf{Geometric Loss.}~
Using $X_{t \to c}$ and $\bm{P}^{\phi}_t$, we compute the \emph{warped} keypoints from image $I_t$ to $I_c$ as:  
\begin{equation} \label{eq:warped_keypoints}
\begin{gathered}
\mathbf{p}^{\phi}_{t\to c} = \pi\left( X_{t \to c}\bm{P}^{\phi}_t\right) = \pi(R\cdot \bm{P}^{\phi}_t + t)
\end{gathered}
\end{equation}
At training time, we aim to minimize the distance between the set of warped keypoints $\bm{p}^{\phi}_{t \to c}$ and the set of corresponding keypoints $\bm{p}^{\phi}_c$ obtained via descriptor matching (Eq.~\ref{eq:descriptor_matching}):
\begin{equation} \label{eq:geometric_loss}
\begin{gathered}
\mathcal{L}_{geom} = || \bm{p}^{\phi}_c - \bm{p}^{\phi}_{t \to c} ||_2
\end{gathered}
\end{equation}

\textbf{Descriptor Loss.}~
Following~\citet{tang2019neural}, we use nested hardest sample mining to self-supervise the keypoint descriptors between the two views. Given \textit{anchor} descriptors $\bm{f}_t$ from the target frame and their associated \textit{positive} descriptors $\bm{f}_{+}=\bm{f}^{\phi}_t$ in the source frame, we define the triplet loss:
\begin{equation} \label{eq:descriptor_loss}
\begin{gathered}
\mathcal{L}_{desc} = \max (0, \|\bm{f}, \bm{f}_{+}\|_2 - \|\bm{f}, \bm{f}_{-}\|_2 + m)~,
\end{gathered}
\end{equation}
where $\bm{f}_{-}$ is the hardest descriptor sample mined from $\bm{f}_s$ with margin $m$.

\textbf{Score Loss.}~
The score loss is introduced to identify reliable and repeatable keypoints in the matching process. In particular, we want to ensure that (i) the feature-pairs have consistent scores across matching views; and (ii) the network learns to predict high scores for good keypoints with low geometric error and strong repeatability. Following~\citet{christiansen2019unsuperpoint}, we achieve this by minimizing the squared distance between scores for each matched keypoint-pair, and minimizing (respectively maximizing) the average score of a matched keypoint-pair if the distance between the paired keypoints is greater (respectively smaller) than the average distance: 
\begin{equation} \label{eq:score_loss}
\begin{gathered}
\mathcal{L}_{score} = \left[ \frac{(\bm{s}^{\phi}_t + \bm{s}^{\phi}_c)}{2}\cdot (\|(\bm{p}^{\phi}_{t \to c},\bm{p}^{\phi}_c\|_2- \bar{\bm{d}}) + (\bm{s}^{\phi}_t -\bm{s}^{\phi}_c)^2\right]~,
\end{gathered}
\end{equation}
where $\bm{s}^{\phi}_t$ and $\bm{s}^{\phi}_c$ are the scores of the target and context keypoints respectively, and $\bar{\bm{d}}$ is the average reprojection error of corresponding keypoints given by $\bar{\bm{d}}=\sum_{i}^{L}\frac{\|\left(p^{\phi}_{i \to c}\right)^2 - \left(p^{\phi}_i\right)^2\|_2}{L}$, and $L$ denotes the total number of keypoint pairs.

\subsubsection{Depth Loss}

The total depth loss is also composed of three terms: $\mathcal{L}_{depth} = \mathcal{L}_{photo} + \beta_3\mathcal{L}_{smooth} + \beta_4\mathcal{L}_{const}$.

\textbf{Photometric loss.} 
Following~\citep{zhou2017unsupervised,godard2018digging,guizilini2019packnet}, we warp the estimated dense depth of the target image $D_t$ via the predicted ego-motion estimate $X_{t \to c}$ to the context frame $I_c$. Using~\citep{jaderberg2015spatial}, we synthesize $\hat{I}_t = I_c\left(q_{t\to c} \right)$ for all pixels $q_t \in I_t$, where $q_{t \to c}$ is the context pixel computed by warping the target pixel $q_t$ via Equation \ref{eq:warped_keypoints}. This operation is done via grid sampling with bilinear interpolation~\citep{jaderberg2015spatial} and thus is differentiable. 
Following~\citep{zhou2017unsupervised,godard2018digging,jaderberg2015spatial}, we impose a dense photometric loss which consists of a structural similarity (SSIM) loss~\citep{wang2004image} (defined in Appendix~\ref{appendix:sec:SSIM}) and an L1 pixel-wise loss term:
\begin{equation} \label{eq:photo_loss}
\begin{gathered}
 \mathcal{L}_{photo}\left(I_t,\hat{I_t}\right) = \gamma~\frac{1 - \text{SSIM}\left(I_t,\hat{I_t}\right)}{2} + \left(1-\gamma\right)~| I_t - \hat{I_t} |~.
\end{gathered}
\end{equation}
In addition, we mask out static pixels which have a \textit{warped} photometric loss $\mathcal{L}_{photo}(I_t, \hat{I}_t)$ higher than their corresponding \textit{unwarped} photometric loss $\mathcal{L}_{photo}(I_t, I_{c})$, calculated using the original source image without view-synthesis as described in~\citep{godard2018digging}. Additionally, we employ a smoothness loss term $\mathcal{L}_{smooth}$ following~\citep{godard2017unsupervised} which we describe in detail in Appendix~\ref{appendinx:sec:depth_losses}. 

\textbf{Depth Consistency.}~
While recovering scale-consistent depth is not a strict requirement for the proposed framework to learn 3D keypoints, scale-consistency is important for tasks that involve accurate ego-motion estimation~\cite{bian2019unsupervised,gordon2019depth}. To this end, we incorporate a depth consistency term that discourages scale-drift between dense depth predictions in adjacent frames:
\begin{equation}\label{eq:consistency_loss}
\begin{gathered}
\mathcal{L}_{const} = \frac{\|D_t(\bm{p}^{\phi}_t) - D_c(\bm{p}^{\phi}_c)\|}{D_t(\bm{p}^{\phi}_t) + D_c(\bm{p}^{\phi}_t)}
\end{gathered}
\end{equation}
Note that $\mathcal{L}_{const}$ is a sparse loss defined based on the correspondences $\left[\mathbf{p}^{\phi}_t, \mathbf{p}^{\phi}_c\right]$.

%% file: latex/040experiments.tex
\section{Experiments}
\label{sec:experiments}

We evaluate our system on the \textit{KITTI}~\citep{geiger2013vision} dataset and report $t_{rel}$ - average translational RMSE drift (\%) on trajectories of length 100-800m, and $r_{rel}$ - average rotational RMSE drift (deg /100m) on trajectories of length 100-800m. All our $t_{rel}$ results are obtained after performing a single Sim(3) alignment step~\citep{grupp2017evo} w.r.t. the ground truth trajectories. We report results for the two main protocols used in related works. Most end-to-end learning based methods train on the KITTI sequences 00-08 and evaluate on seq. 09 and 10. These results are summarized in Table~\ref{table:kitti-odom-results-horizontal}, with more details in Appendices~\ref{appendix:sec:pose_ablation} and~\ref{appendix:sec:qualitative_odometry}. A number of methods including DVSO~\citep{yang2018deep} and D3VO~\citep{yang2020d3vo} train on the \textit{Eigen}~\citep{eigen2014depth} train split, which includes sequences 01, 02, 06, 08, 09 and 10; in this case the test sequences are 00, 03, 04, 05 and 07. We summarize our results for this protocol in Table~\ref{table:kitti-eigen-traj}. Additionally, we report in the standard depth evaluation metrics on the \textit{Eigen}~\citep{eigen2014depth} test split in Appendix~\ref{appendix:sec:depth_evaluation}. 

To evaluate the performance of our kepoint detector and decriptor we use the \textit{HPatches}~\citep{balntas2017hpatches} dataset, which contains a set of 116 image sequences (illumination and viewpoint) for a total of 580 image pairs. We quantify detector performance through the \textit{Repeatability} and \textit{Localization Error} metrics and descriptor performance through the \textit{Correctness} and \textit{Matching Score} metrics (the exact definition of these metrics can be found in Appendix~\ref{appendix:keypoint_metrics}). For a fair comparison, we evaluate the results generated without applying Non-Maxima Suppression (NMS). Following related work~\citep{tang2019neural,christiansen2019unsuperpoint,detone2018superpoint}, we pre-train our \textit{KeypointNet} on the \textit{COCO}~\citep{lin2014microsoft} dataset, which contains $118k$ training images. We note that pretraining on COCO is self-supervised, as described in Sec.~\ref{subsec:training}. 

\vspace{-2mm}
\subsection{Training}
\label{subsec:training}

We implement our networks in PyTorch~\citep{paszke2017automatic} using the ADAM optimizer~\citep{kingma2014adam}. We use $10^{-4}$ as the learning rate and train KeypointNet and DepthNet jointly for $50$ epochs with a batch size of $8$. We implement KeypointNet following~\citep{tang2019neural}, with the mention that we use an ImageNet pre-trained ResNet-18 backbone, which we find performs better than the reference architecture. We follow~\citep{godard2018digging} and implement DepthNet using an ImageNet~\citep{Deng09imagenet} pretrained ResNet-18 backbone along with a depth decoder that outputs inverse depth at 4 different resolution scales. However, at test-time, only the highest resolution scale is used for 2D-to-3D keypoint lifting. We describe our networks in detail in Appendix~\ref{appendix:network_diagrams}. We train on snippets of 3 images $\left(I_{t-\Delta t}, I_t, I_{t+ \Delta t}\right)$, for $\Delta t\in[1,2,4]$ with target image $I_t$ and images $\left(I_{t-\Delta t}, I_{t+\Delta t}\right)\in I_C$ as context images. Using the pair of target and context images we compute the losses as defined in Section~\ref{subsec:learning}. The dense \textit{photometric} loss is computed over the context $I_C$ as shown in Equation~\ref{eq:photo_loss}. In all our experiments we set $\alpha=0.1$, $\beta_1=\beta_2=1.0$, $\beta_3=\beta_4=0.1$, and $\gamma=0.85$. Additionally, starting from the target image $I_t$, we also perform Homography Adaptation similar to~\citep{tang2019neural}, e.g., translation, rotation, scaling, cropping and symmetric perspective transform and we apply per-pixel Gaussian noise, color jitter, and Gaussian blur to the images. \rares{Our method runs at 30fps on images of resolution 640x192 on a V100 GPU}.

\textbf{Pretrained baseline.}~We pretrain KeypointNet on \textit{COCO} using Homography Adaptation for $50$ epochs using a learning rate of $5 \cdot 10^{-4}$ which is halved after $40$ epochs. We refer to this as our baseline KeypointNet, and evaluate its performance in Table~\ref{table:results_descriptor}. To speed up convergence, we pretrain our DepthNet on the KITTI training sequences using the method described in~\citep{godard2018digging}. We train for 200 epochs with a learning rate of $10^{-4}$ which is decayed by $0.5$ every $40$ epochs. We refer to this as our baseline DepthNet, and we evaluate its performance in the ablative evaluations in Table~\ref{table:ablative}.
\input{latex/041dso}

%% file: latex/041dso.tex
\vspace{-2mm}
\subsection{Direct Sparse Odometry (DSO) Integration}
\label{sec:ds-dso}

To evaluate tracking performance, we integrate our self-supervised, depth-aware keypoints into the DSO framework of~\citet{engel2017direct}, and we show that we are able to achieve long-term tracking results which are especially on par with stereo methods such as DVSO~\citep{yang2018deep} or ORB-SLAM2~\citep{mur2017orb}. Unlike other monocular visual odometry approaches, the superior keypoint matching and stable 3D lifting performance of our proposed method allows us to bootstrap the tracking system, rejecting false matches and outliers and avoiding significant scale-drift. Our integration is built on top of the windowed sparse direct bundle adjustment formulation of DSO. Specifically, we improve depth-initialization of keyframes in the original DSO implmenetation by using the depth estimated through our proposed self-supervised 3D keypoints. In addition, we modify the hand-engineered direct semi-dense tracking component to use the proposed sparse and robust learned keypoint-based method introduced in this work. We describe the DSO integration in more details in Appendix~\ref{appendix:sec:ds-dso}.

%% file: latex/05results.tex
\subsection{Visual Odometry Performance}
\label{subsec:vo-results}

\input{latex/results/kitti_eigen_traj_results.tex}

\rares{We summarize our visual odometry results and comparisons with state-of-the-art methods in Table~\ref{table:kitti-eigen-traj}, and we note that our method outperforms all other monocular-trained methods, particularly in the $t_{rel}$ metric. For $r_{rel}$ we note that~\citep{ambrus2019two} achieves better results on some sequences, which we attribute to difficulty in matching keypoints in those settings, however our $r_{rel}$ is superior on average over the test sequences.} Our method also outperforms stereo-trained methods except for DVSO~\citep{yang2018deep} and D3VO~\citep{yang2020d3vo}. However, we emphasize that while both~\citep{yang2018deep,yang2020d3vo} are trained from a wide-baseline stereo setup which provides a strong prior for outlier rejection, our system is trained in a self-supervised manner purely relying on monocular videos - a significantly harder problem. The results indicate that when integrated with tracking frameworks such as DSO our depth-aware keypoints provide superior matching performance that even rivals state-of-the-art methods trained on stereo imagery. 

\input{latex/results/kitti_odom_results.tex}

Additionally, we report our results when training on KITTI sequences 00-08 in  Table~\ref{table:kitti-odom-results-horizontal}. We note that our method outperforms all other monocular methods except for DF-VO~\citep{zhan2019visual} and~\citep{zhao2020towards} for $r_{rel}$. Both~\citep{zhao2020towards,zhan2019visual} heavily rely on optical-flow and RANSAC-based essential/fundamental matrix computation and scale-factor recovery. Recall that our method uses PnP (as described in Section~\ref{subsec:pnp_pose}) for pose estimation. When comparing with the PnP-based version DF-VO~\citep{zhan2019visual} (DF-VO PnP in Table~\ref{table:kitti-odom-results-horizontal}), we observe that our method performs much better, which we attribute to the direct optimization of sparse 2D-3D keypoints, as opposed to~\citep{zhan2019visual} which relies purely on dense optical flow. \rares{Finally, we record that our method performs worse on sequence 01 compared to the other sequences. In this sequence the vehicle is driving at high speeds on the highway and the image contains a high number of sky pixels, making this a challenging environment for feature detection and matching; we note that a number of other methods also perform poorly in this setting~\citep{ambrus2019two,li2017undeepvo,pillai2018superdepth,zhan2019visual}.}

\subsection{Ablation Study}
\label{subsec:ablative_study}
We summarize our ablative analysis in Table~\ref{table:ablative}. Our baseline - KeypointNet pre-trained on COCO and DepthNet trained on KITTI, but the two are not optimized together - shows superior results compared to most monocular methods (see Tables~\ref{table:kitti-eigen-traj} and~\ref{table:kitti-odom-results-horizontal}), thus motivating our approach of combining keypoints and depth in a self-supervised learning framework.  We notice a significant improvement when training the two networks together (\textit{Row 2: Ours $-$ Diff Pose}). Adding the differential pose estimation (\textit{Row 5: Ours }) further improves the performance of our system for the $t_{rel}$ metric; we note that the $r_{rel}$ test metric does not improve, mostly due to an error in Sequence $04$ (please refer to Appendix~\ref{appendix:sec:pose_ablation} for detailed results). We further ablate the KeypointNet (\textit{Row 3: Ours $-$ KPN trained}) - i.e., we estimate the ego-motion using the DepthNet after training together with the KeypointNet, but we use the original KeypointNet trained only on COCO. We perform a similar experiment ablating the trained DepthNet (\textit{Row 4: Ours $-$ DN trained}). Finally, when integrated in DSO (\textit{Row 6: Ours DSO}) our results improve significantly, which we attribute to the robustness of our features to scene appearance and geometry. We provide more details in Appendix~\ref{appendix:sec:pose_ablation}, and emphasize that all our experiments, including pretraining, are done in a self-supervised fashion, without any supervision.

\input{latex/results/KITTI_traj_ablation.tex}

\subsection{Keypoint Detector and Descriptor Performance}
\label{subsec:keypoint-results}

Table~\ref{table:results_descriptor} shows the performance of our method on HPatches~\citep{balntas2017hpatches}. We note that by improving the network capacity and using a \textit{ResNet-18} architecture, our baseline outperforms all classical as well as learning-based methods (recall that our KeypointNet baseline is trained only on COCO, and can thus be directly compared to these methods). Further, we note similar metrics between our baseline method and our method after training on KITTI (\textit{KeypointNet baseline vs KeypointNet}). We conclude that by training with the proposed losses in a joint self-supervised framework, our keypoints have gained geometric understanding and can be used for state-of-the-art ego-motion estimation while at the same time retaining their robustness to illumination and viewpoint changes.

\input{latex/results/hpatches.tex}

%% file: latex/results/kitti_eigen_traj_results.tex
\begin{table}[!t]
\centering
\small
\renewcommand{\b}[1]{\textbf{#1}}
\renewcommand{\u}[1]{\underline{#1}}
\setlength{\tabcolsep}{.6em}

\resizebox{0.98\hsize}{!}{
\begin{tabular}{lcccccccccccccc}

\textbf{Method} & {Type} & 01 &    02  &  06 &  08  &   09 & 10 & 00  & 03  & 04  & 05  & 07 & {Train} & {Test} \\
\toprule

& \multicolumn{12}{c}{$t_{rel}$ - Average Translational RMSE drift (\%) on trajectories of length 100-800m.}  \\
\midrule
\rares{Mono-DSO~\cite{engel2014lsd}}
& Mono   & 9.17   &  114  &  42.2 &  177 &  28.1  & 24.0 &  -   & -  & -  & -  & - &  65.75 & - 
\\
ORB-SLAM-M~\cite{mur2017orb}                
& Mono   &  -  &  -  &  -  &  32.40 &  -  &  -  &  25.29  & -  &  -  &  26.01 & 24.53  & -  & 27.05 
\\
Ambrus et al~\cite{ambrus2019two}                
& Mono   & 17.59 &  6.82 & 8.93 &  8.38 &  6.49 & 9.83 &  7.16 &  7.66 & 3.8   &   6.6 &  11.48 & 9.67 &  7.34 
\\
UnDeepVO~\cite{li2017undeepvo}                   
& Stereo & 69.1 & 5.58 &  6.20 &  4.08 & 7.01 & 10.6 & 4.14 & 5.00 & 4.49 & 3.40 &  3.15 & 11.68 & 8.81
\\
SuperDepth~\cite{pillai2018superdepth} 
& Stereo & 13.48 & 3.48 & 1.81  & 2.25  &  3.74 &  2.26 & 6.12  & 7.90  & 11.80 &  4.58 & 7.60  & 4.50  & 7.60  
\\
DVSO~\cite{yang2018deep} 
& Stereo & 1.18 &  0.84 & 0.71 &  1.03 &  0.83 &  0.74 & \u{0.71}  &  \u{0.77} & \u{0.35} &  \u{0.58} &  0.73 & \u{0.89}  & \u{0.63}
\\
D3VO~\cite{yang2020d3vo}
& Stereo & \u{1.07} & \u{0.80} & \u{0.67} & \u{0.78} & \u{0.62} & - & - & - & - & - & - & - & 0.82
\\
\midrule
Ours
& Mono & 17.79 & \b{3.15} & 1.88 &  3.06 &  2.69 & \b{5.12} & 2.76 &  3.02 & 1.93 & 3.30 & 2.41 & 5.61 & 2.68 
\\
Ours + DSO                             
& Mono  & \b{4.70} & 3.62 & \b{0.92} &  \b{2.46} &  \b{2.31} &  5.24 & \b{1.83} &  \b{1.21} & \b{0.76} & \b{1.84} & \b{\u{0.54}} & \b{3.21} &  \b{1.24}
\\
\midrule
& \multicolumn{12}{c}{$r_{rel}$ - Average Rotational RMSE drift ($^{{\circ}}/100m$) on trajectories of length 100-800m.} \\ 
\midrule
ORB-SLAM-M~\cite{mur2017orb}  
& Mono  &  -  &  -  &  -  &  12.13 &  -  &  -  &  7.37  & -  &  -  &   10.62 & 10.83  & -  & 10.23 
\\
Ambrus et al~\cite{ambrus2019two}             
& Mono   & 1.01 & 0.87 & 0.39 & 0.61 & 0.86 & 0.98 & 1.70 & 3.49 & 0.42 & 0.90 & 2.05 & 0.79 & 1.71
\\
UnDeepVO~\cite{li2017undeepvo}                   
& Stereo &  1.60 & 2.44 & 1.98 &  1.79 & 3.61 & 4.65 &  1.92 & 6.17 & 2.13 & 1.5 &  2.48 &  2.45 & 4.13 
\\
SuperDepth~\cite{pillai2018superdepth}           
& Stereo & 1.97 & 1.10 & 0.78 & 0.84 & 1.19 & 1.03 & 2.72& 4.30  & 1.90 & 1.67 & 5.17 & 1.15 & 3.15 
\\
DVSO~\cite{yang2018deep}                         
& Stereo & \u{0.11} & \u{0.22} & 0.20 & 0.25 & \u{0.21} & \u{0.21} & \u{0.24} & \u{0.18} & \u{0.06} & \u{0.22} & 0.35 & \u{0.20} & \u{0.21}
\\
\midrule
Ours                                       
& Mono  & 0.72 & 1.01 & 0.80 &  0.76 &  0.61 & 1.07 &  1.17 &  2.45 & 1.93 & 1.11 & 1.16 & 0.83 & 1.56 
\\
Ours + DSO
& Mono   & \b{0.16} & \b{\u{0.22}} & \b{\u{0.13}} & 0.31 &  \b{0.30} & \b{0.29} &  \b{0.33} & \b{0.33} & \b{0.18}  & \b{\u{0.22}} &  \b{\u{0.23}} & \b{0.24} &  \b{0.26}
\\
\bottomrule
\end{tabular}
}
\vspace{1mm}
\caption{\textbf{Comparison of vision-based trajectory estimation with state-of-the-art methods.}~The \textit{Type} column indicates the data used at training time. Note: All methods are evaluated on monocular data. \textbf{Bold} text denotes the best method trained on monocular data; \_ denotes the best overall method. Training seq. 01, 02, 06, 08, 09 and 10; test seq. 00, 03, 04, 05 and 07. The numbers for~\cite{mur2017orb} are reported from~\cite{li2017undeepvo}; \rares{the numbers for~\cite{engel2014lsd} are reported from~\cite{yang2020d3vo}}.
}
\label{table:kitti-eigen-traj} 
\end{table}

%% file: latex/results/kitti_odom_results.tex
\begin{table}[!t]
\vspace{-4mm}
\centering
\small
\renewcommand{\b}[1]{\textbf{#1}}
\renewcommand{\u}[1]{\underline{#1}}
\setlength{\tabcolsep}{.6em}

\resizebox{0.98\hsize}{!}{
\begin{tabular}{lccccccccc}

&  Bian et al~\cite{bian2019unsupervised} & EPC++~\cite{luo2019every} & Monodepth2$\ddagger$~\cite{godard2018digging} & DF-VO~\cite{zhan2019visual} PnP & Zhao et al~\cite{zhao2020towards} & SGANVO~\cite{feng2019sganvo} &  Gordon et al~\cite{gordon2019depth} & DF-VO~\cite{zhan2019visual} & Ours \\

\toprule

& \multicolumn{7}{c}{$t_{rel}$ - Average Translational RMSE drift (\%) on trajectories of length 100-800m.}  \\
\midrule

Seq 09 & 11.2  &  8.84 &  5.28 & 7.12 & 6.93 & 4.95 & \u{2.7} & \b{2.47} & 3.14 \\
Seq 10 & 10.1  &  8.86 &  8.47 & 6.83 & \u{4.66} & 5.89 & 6.8 &\b{1.96} & 5.45 \\
Mean   & 10.7  &  8.85 &  7.14 & 6.98 & 5.76 & 5.42 & 4.75 & \b{2.21} & \u{4.30} \\

\midrule
& \multicolumn{7}{c}{$r_{rel}$ - Average Rotational RMSE drift ($^{{\circ}}/100m$) on trajectories of length 100-800m.} \\ 
\midrule

Seq 09 & 3.35 & 3.34 & 1.60 & 2.43 & \u{0.44} & 2.37 & - & \b{0.30} & 0.73 \\
Seq 10 & 4.96 & 3.18 & 2.26 & 3.88 & \u{0.62} & 3.56 & - & \b{0.31} & 1.18 \\
Mean   & 4.20 & 3.26 & 1.73 & 3.12 & \u{0.53} & 3.02 & - & \b{0.31} & 0.90  \\

\bottomrule
\end{tabular}}
\vspace{1mm}
\caption{\textbf{Comparison of vision-based trajectory estimation with state-of-the-art monocular methods.}~\textbf{Bold} text denotes the best performing method; \_ denotes the second best method. Training seq. 00-08; test seq: 09 and 10. $\ddagger$ -  the numbers of~\cite{godard2018digging} are based on our own implementation.
}
\label{table:kitti-odom-results-horizontal} 
\vspace{-4mm}
\end{table}

%% file: latex/results/KITTI_traj_ablation.tex
\begin{table}[!t]
\centering
{
\renewcommand{\b}[1]{\textbf{#1}}
\renewcommand{\b}[1]{\textbf{#1}}
\renewcommand{\r}[1]{\rotatebox{90}{#1}}
\newcommand{\ch}{$\checkmark$}
\setlength{\tabcolsep}{.9em}
\resizebox{0.98\hsize}{!}{
\begin{tabular}{l|cccccc|cccc|}
& {KPN} & {{DN}} & {{KPN}} & {{DN}} & {{Diff}} & \multirow{2}{*}{DSO} & \multicolumn{2}{c}{$r_{rel}$} & \multicolumn{2}{c|}{$t_{rel}$} \\
\textbf{Method} & {baseline} & {{baseline}} & {{trained}} & {{trained}} & {{Pose}} &  & train & test & train & test 
\\
\toprule
1. Baseline
& \ch & \ch & - & - & - & -  & 1.02 & 1.63 & 6.08 & 3.14 
\\
2. Ours $-$ Diff Pose
& \ch & \ch & \ch & \ch & - & -  & 0.89 & 1.43 & 6.12 & 2.92 
\\
3. Ours $-$ KPN trained
& \ch & \ch & - & \ch & \ch & - & 0.93 & 1.61 & 5.94 & 2.88  
\\
4. Ours $-$ DN trained
& \ch & \ch & \ch & - & \ch & -  & 0.91 & 1.58 & 5.38 & 2.88 
\\
5. Ours
& \ch & \ch & \ch & \ch & \ch & -   & 0.83 & 1.56  & 5.61 & 2.68
\\
6. Ours + DSO
& \ch & \ch & \ch & \ch & \ch & \ch & 0.24 & 0.26 & 3.21 & 1.24  
\\
\bottomrule                     
\end{tabular}
}
}
\vspace{1mm}
\caption{\textbf{Ablative analysis}: \textbf{Diff Pose} - the differentiable pose component (Section~\ref{subsec:pnp_pose}), \textbf{KPN trained} - the trained version of the KeyPointNet (i.e., using only the baseline), \textbf{DN trained} - the trained version of the DepthNet. Training seq.  01, 02, 06, 08, 09, 10; test seq.  00, 03, 04, 05, 07.}
\label{table:ablative}
\end{table}

%% file: latex/results/hpatches.tex
\begin{table}[!t]
\vspace{-4mm}
\centering
{
\renewcommand{\b}[1]{\textbf{#1}}
\tiny
\setlength{\tabcolsep}{.7em}
\resizebox{0.98\hsize}{!}{
\begin{tabular}{l|c|c|ccc|c||c|c|ccc|c}
  \multirow{ 2}{*}{\textbf{Method}}                  & \multicolumn{6}{c}{240x320, 300 points}        &    \multicolumn{6}{c}{480 x 640, 1000 points}   \\ 
{}             & Rep. & Loc. & Cor-1   & Cor-3 & Cor-5 & M.Score & Rep. & Loc. & Cor-1 & Cor-3 & Cor-5 & M.Score    \\  
\toprule

ORB~\cite{rublee2011orb}                
                  & 0.532 & 1.429 & 0.131 & 0.422 & 0.540 & 0.218 & 
                     0.525 & 1.430 & 0.286 & 0.607 & 0.71  & 0.204 \\
SURF~\cite{bay2006surf}
                  & 0.491 & 1.150 & 0.397 & 0.702 & 0.762 & 0.255 & 
                     0.468 & 1.244 & 0.421 & 0.745 & 0.812 & 0.230 \\
BRISK~\cite{leutenegger2011brisk}
                  & 0.566 & 1.077 & 0.414 & 0.767 & 0.826 & 0.258 & 
                     0.746 & 0.211 & 0.505 & 1.207 & 0.300 & 0.653 \\
SIFT~\cite{lowe1999object}
                  & 0.451 & 0.855 & 0.622 & 0.845 & 0.878 & 0.304 & 
                     0.421 & 1.011 & 0.602 & 0.833 & 0.876 & 0.265 \\
LF-Net(indoor)~\cite{ono2018lf}
                  & 0.486 & 1.341 & 0.183 & 0.628 & 0.779 & 0.326 & 
                     0.467 & 1.385 & 0.231 & 0.679 & 0.803 & 0.287 \\
LF-Net(outdoor)~\cite{ono2018lf}    
                  & 0.538 & 1.084 & 0.347 & 0.728 & 0.831 & 0.296 & 
                     0.523 & 1.183 & 0.400 & 0.745 & 0.834 & 0.241 \\
SuperPoint~\cite{detone2018superpoint}         
                  & 0.631 & 1.109 & 0.491 & 0.833 & 0.893 & 0.318 & 
                     0.593 & 1.212 & 0.509 & 0.834 & 0.900 & 0.281 \\
UnsuperPoint~\cite{christiansen2019unsuperpoint}       
                  & 0.645 & 0.832 & 0.579 & 0.855 & 0.903 & 0.424 & 
                     0.612 & 0.991 & 0.493 & 0.843 & 0.905 & 0.383 \\     
\hline
 \hline
IO-Net~\cite{tang2019neural}
                  & \b{0.686} & 0.890 & 0.591 & 0.867 & 0.912 & 0.544 & 
                     \b{0.684} & 0.970 & 0.564 & 0.851 & 0.907 & 0.510 \\
KeypointNet Baseline         
                  & 0.683 & 0.816 & \b{0.624} & \b{0.879} & \b{0.924} & 0.573 & 
                     0.682 & 0.898 & \b{0.581} & 0.848 & 0.913 & \b{0.534} \\
KeypointNet        & \b{0.686} & \b{0.799} & 0.532 & 0.858 & 0.906 & \b{0.578} & 
                     0.674 & \b{0.886} & 0.529 & \b{0.867} & \b{0.920} & 0.529 \\
\bottomrule                     
\end{tabular}
}
}
\vspace{1mm}
\caption{\textbf{Keypoint and descriptor performance on HPatches}~\cite{balntas2017hpatches}. Repeatability and Localization Error measure keypoint performance while Correctness (pixel threshold 3) and Matching score measure descriptor performance. Higher is better for all metrics except Localization Error.}
\label{table:results_descriptor}
\vspace{-6mm}
\end{table}

%% file: latex/06conclusion.tex
\vspace{-2mm}
\section{Conclusion}
\label{sec:conclusion}	
\vspace{-2mm}
In this paper, we proposed a fully self-supervised framework for depth-aware keypoint learning from unlabeled monocular videos by incorporating a novel differentiable pose estimation module that simultaneously optimizes the keypoints and their depths in a Structure-from-Motion setting.  
Unlike existing learned keypoint methods that employ only homography adaptation, we exploit the temporal context in videos to further boost the repeatability and matching performance of our proposed keypoint network. The resulting 3D keypoints and associated descriptors exhibit superior performance compared to all other traditional and learned methods, and are also able to learn from realistic non-planar 3D scenes.
Finally, we show how our proposed network can be integrated with a monocular visual odometry system to achieve accurate, scale-aware, long-term tracking results that are on par with state-of-the-art stereo-methods.

%% file: latex/070suppmat.tex
\section{Architecture Diagram} 
\label{appendix:network_diagrams}
\textbf{ResNet18-DepthNet.} we provide a detailed description of our DepthNet architecture in Table~\ref{table:depthnet-arch}. Note that we follow~\citet{godard2017unsupervised} and use a \textit{ResNet18} encoder followed by a decoder which outputs inverse depth at 4 scales.

\begin{table}[!h]
\small
\renewcommand{\arraystretch}{1.1}
\centering
\begin{tabular}{l|l|c|c}
\toprule
& \textbf{Layer Description} & \textbf{K} & \textbf{Output Tensor Dim.} \\ 
\toprule
\#0 & Input RGB image & & 3$\times$H$\times$W \\ 
\midrule
\midrule
\multicolumn{4}{c}{\textbf{ResidualBlock}} \\ \hline
& Conv2d + BatchNorm + ReLU & 3 &  \\
& Conv2d + BatchNorm        & 3 &  \\
\midrule
\multicolumn{4}{c}{\textbf{Depth Encoder}} \\ \hline
\#1  & Conv2d (S2) + BatchNorm + ReLU & 7 &  64$\times$H/2$\times$W/2 \\
\#2  & Conv2d + BatchNorm + ReLU            & 3 &  64$\times$H/2$\times$W/2 \\
\#3  & ResidualBlock (\#2)  x2              & - &  64$\times$H/2$\times$W/2 \\
\#4  & Max. Pooling ($\times$1/2)          & 3 &  64$\times$H/4$\times$W/4 \\
\#5  & ResidualBlock (\#3 + \$2) x2         & - & 128$\times$H/4$\times$W/4 \\
\#6  & Max. Pooling ($\times$1/2)          & 3 & 128$\times$H/8$\times$W/8 \\
\#7  & ResidualBlock (\#4 + \#3) x2         & - & 256$\times$H/8$\times$W/8 \\
\#8  & Max. Pooling ($\times$1/2)          & 3 & 256$\times$H/16$\times$W/16 \\
\#9  & ResidualBlock (\#5 + \#4) x2         & - & 512$\times$H/16$\times$W/16 \\
\midrule
\multicolumn{4}{c}{\textbf{Depth Decoder}} \\ 
\midrule
\#10 & Conv2D + ELU (\#9)                                   & 3 & 128$\times$H/16$\times$W/16 \\
\#11 & Conv2D + Upsample (\#10)                             & 3 & 128$\times$H/8$\times$W/8  \\
\textbf{\#12} & Conv2D + Sigmoid                            & 3 & 1$\times$H/8$\times$W/8 \\
\#13 & Conv2D + ELU                                         & 3 & 64$\times$H/8$\times$W/8 \\
\#14 & Conv2D + Upsample(\#7 $\oplus$ \#13)                 & 3 & 64$\times$H/4$\times$W/4  \\
\textbf{\#15} & Conv2D + Sigmoid                            & 3 & 1$\times$H/8$\times$W/8 \\
\#16 & Conv2D + ELU                                         & 3 & 32$\times$H/4$\times$W/4 \\
\#17 & Conv2D + Upsample (\#5 $\oplus$ \#16)                & 3 & 32$\times$H/2$\times$W/2  \\
\textbf{\#18} & Conv2D + Sigmoid                            & 3 & 1$\times$H/8$\times$W/8 \\
\#19 & Conv2D + ELU                                         & 3 & 16$\times$H/2$\times$W/2 \\
\#20 & Conv2D + Upsample (\#3 $\oplus$ \#19)                & 3 & 16$\times$H$\times$W  \\
\textbf{\#21} & Conv2D + Sigmoid                            & 3 & 1$\times$H$\times$W \\
\bottomrule
\end{tabular}
\caption{DepthNet diagram. Line numbers in bold indicate output inverse depth layer scales. \emph{Upsample} is a nearest-neighbor interpolation operation that doubles the spatial dimensions of the input tensor. $\oplus$ denotes feature concatenation for skip connections.}
\label{table:depthnet-arch}
\end{table}

\textbf{ResNet18 KeypointNet.} Table~\ref{table:keypointnet_arch} details the network architecture of our KeypointNet. We follow~\cite{tang2019neural} but change the network encoder and use a \textit{ResNet18} architecture instead, which we found to perform better. 

\begin{table}[!h]
\small
\renewcommand{\arraystretch}{1.1}
\centering
\begin{tabular}{l|l|c|c}
\toprule
& \textbf{Layer Description} & \textbf{K} & \textbf{Output Tensor Dim.} \\ 
\toprule
\#0 & Input RGB image & & 3$\times$H$\times$W \\ 
\midrule
\midrule
\multicolumn{4}{c}{\textbf{ResidualBlock}} \\ \hline
& Conv2d + BatchNorm + ReLU & 3 &  \\
& Conv2d + BatchNorm        & 3 &  \\
\midrule
\multicolumn{4}{c}{\textbf{KeyPoint Encoder}} \\ \hline
\#1  & Conv2d (S2) + BatchNorm + ReLU & 7 &  64$\times$H/2$\times$W/2 \\
\#2  & Conv2d + BatchNorm + ReLU            & 3 &  64$\times$H/2$\times$W/2 \\
\#3  & ResidualBlock (\#2)  x2              & - &  64$\times$H/2$\times$W/2 \\
\#4  & Max. Pooling ($\times$1/2)          & 3 &  64$\times$H/4$\times$W/4 \\
\#5  & ResidualBlock (\#3 + \$2) x2         & - & 128$\times$H/4$\times$W/4 \\
\#6  & Max. Pooling ($\times$1/2)          & 3 & 128$\times$H/8$\times$W/8 \\
\#7  & ResidualBlock (\#4 + \#3) x2         & - & 256$\times$H/8$\times$W/8 \\
\#8  & Max. Pooling ($\times$1/2)          & 3 & 256$\times$H/16$\times$W/16 \\
\#9  & ResidualBlock (\#5 + \#4) x2         & - & 512$\times$H/16$\times$W/16 \\
\midrule
\multicolumn{4}{c}{\textbf{KeyPoint Decoder}} \\ 
\midrule
\#10 & Conv2D + BatchNorm + LReLU (\#9)                     & 3 & 256$\times$H/16$\times$W/16 \\
\#11 & Conv2D + Upsample (\#10)                             & 3 & 256$\times$H/8$\times$W/8  \\
\#12 & Conv2D + BatchNorm + LReLU                           & 3 & 256$\times$H/8$\times$W/8 \\
\#13 & Conv2D + Upsample(\#7 $\oplus$ \#12)                 & 3 & 128$\times$H/4$\times$W/4  \\
\#14 & Conv2D + BatchNorm + LReLU                           & 3 & 128$\times$H/4$\times$W/4 \\
\#15 & Conv2D + Upsample (\#5 $\oplus$ \#14)                & 3 & 64$\times$H/2$\times$W/2  \\
\#16 & Conv2D + BatchNorm + LReLU                           & 3 & 64$\times$H/2$\times$W/2 \\
\midrule
\midrule
\multicolumn{4}{c}{\textbf{Score Head}} \\
\midrule
\#12 & Conv2d + BatchNorm + LReLU (\#12) & 3 & 256$\times$H/8$\times$W/8 \\
\#13 & Conv2d + Sigmoid                    & 3 & 1$\times$H/8$\times$W/8 \\
\midrule
\multicolumn{4}{c}{\textbf{Location Head}} \\ 
\midrule
\#14 & Conv2d + BatchNorm + LReLU (\#12) & 3 & 256$\times$H/8$\times$W/8 \\
\#15 & Conv2d + Tan. Harmonic              & 3 & 2$\times$H/8$\times$W/8 \\
\midrule
\multicolumn{4}{c}{\textbf{Descriptor Head}} \\ 
\midrule
\#16  & Conv2d + BatchNorm + LReLU (\#16) & 3 & 64$\times$H/2$\times$W/2 \\
\#17  & Conv2d                          & 3 & 64$\times$H/2$\times$W/2 \\
\bottomrule
\end{tabular}
\caption{KeypointNet diagram. \emph{Upsample} is a nearest-neighbor interpolation operation that doubles the spatial dimensions of the input tensor. $\oplus$ denotes feature concatenation for skip connections.}
\label{table:keypointnet_arch}
\end{table}

\section{Video}
\label{sec:video}

The accompanying video presents our contributions and demonstrates the long-term visual odometry accuracy obtained by our method on a video sequence. The main panel (top, center) shows our real-time semi-dense reconstruction along with the vehicle trajectory. The bottom panel shows (from left to right): the flow of inlier keypoints, color-coded based on depth; the estimated monocular depth; and finally, the number of matched keypoints versus the number tracked inliers. 

\section{Dense Depth Evaluation}
\label{appendix:sec:depth_evaluation}
\input{latex/results/depth_results.tex}

We perform a qualitative evaluation of our DepthNet on the KITTI dataset, specifically on the Eigen~\cite{eigen2014depth} test split, and report the numbers in Table~\ref{table:depth-accuracy}. We also include the numbers reported by~\cite{godard2018digging} and note that our \textit{DepthNet baseline} numbers are on par with those of~\cite{godard2018digging} (which correspond to row \textit{1. Baseline} of Table 3 in the main text). Table~\ref{table:depth-accuracy} also shows our numbers after fine-tuning the DepthNet and KeypointNet through the proposed \textit{Multi-View Adaptation} method (which correspond to row \textit{5. Ours} of Table 3 in the main text). We note a slight decrease in the \textit{Abs Rel} and \textit{Sq Rel} metrics, but otherwise the numbers are within error margin with respect to our baseline. These results provide an important sanity check: as the main focus of this work is sparse, depth-aware keypoint learning, we don't expect to see much variation when performing dense depth evaluation. We mention that sparsely evaluating the depth using the keypoints regressed by our method is not feasible using the depth available in the KITTI dataset: even using the denser depth maps provided by~\cite{uhrig2017sparsity}, only about $10\%$ of our keypoints have valid depths in the ground truth maps, which amounts to a very small number of points ($<50$) per image.  

\input{latex/073suppmat_dso}

\section{Depth losses}
\label{appendinx:sec:depth_losses}
\textbf{Depth Smoothness Loss.}~
In order to regularize the depth in texture-less low-image gradient regions, we use an edge-aware loss term similar to~\cite{godard2017unsupervised}: 
\begin{align}
\mathcal{L}_{smooth} = | \delta_x \hat{D}_t | e^{-|\delta_x I_t|} + | \delta_y \hat{D}_t | e^{-|\delta_y I_t|}~.
\label{eq:loss-disp-smoothness}
\end{align}

\textbf{Structural Similarity (SSIM) loss.}
\label{appendix:sec:SSIM}
\input{latex/071suppmat_ssim.tex}

\section{Pose Estimation}
\label{appendix:sec:pose_pnp}
\input{latex/072suppmat_pnp_pose.tex}

\section{Detailed Results for the Pose Ablation Study }
\label{appendix:sec:pose_ablation}

Table~\ref{table:kitti-eigen-traj-appendix} provides detailed results on all the KITTI odometry sequences for each entry  of our ablation study (Table 3 of the main text). We note that (i) the proposed contributions - i.e. row 2 vs row 1 and row 5 vs row 1 consistently improve over the baseline; and that (ii) by swapping out the KeypointNet or DepthNet trained using the proposed method with their baseline counterparts (rows 3 and 4) results in worse performance for both the $t_{rel}$ and $r_{rel}$ metrics. Thus we conclude that the proposed training procedure along with the differentiable pose component improves both the DepthNet and KeypointNet for the task of Visual Odometry. 



\input{latex/results/appendix_traj_ablation_results.tex}

\section{Additional KITTI odometry results}
\label{appendix:sec:pose_kitti_odom}

We show additional comparisons with state-of-the-art methods in Table~\ref{table:appendix-kitti-odom-results-horizontal}, complementing our results from Table 2 of the main text. Additionally, we present detailed results of our method on all the KITTI odometry sequences in Table~\ref{table:supp-kitti-odom-results-ours-detailed}.

\input{latex/results/supp_kitti_odom_results}

\input{latex/results/supp_kitti_odom_ours_detailed}

\section{Qualitative results on KITTI odometry sequences}
\label{appendix:sec:qualitative_odometry}

\begin{figure}[!t]
\centering
\resizebox{\hsize}{!}{
	\includegraphics[width=0.45\columnwidth]{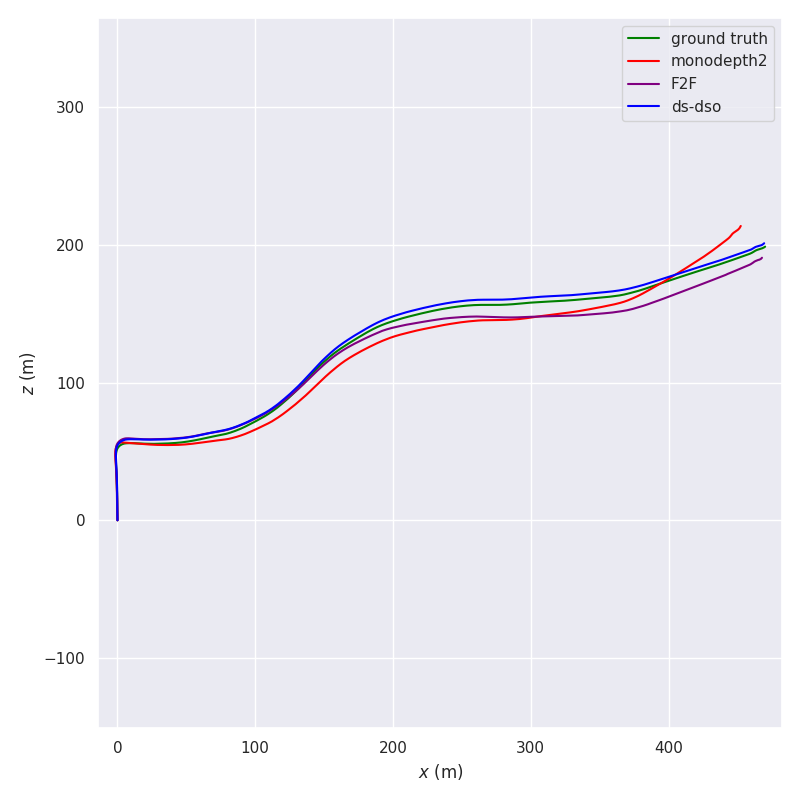} \includegraphics[width=0.45\columnwidth]{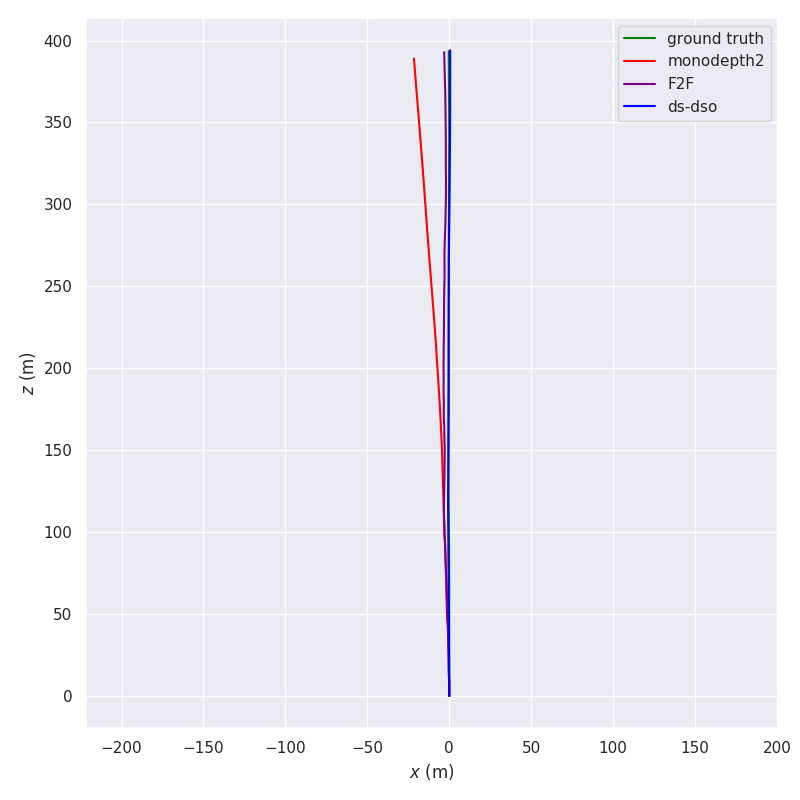}}
\resizebox{\hsize}{!}{
    \includegraphics[width=0.45\columnwidth]{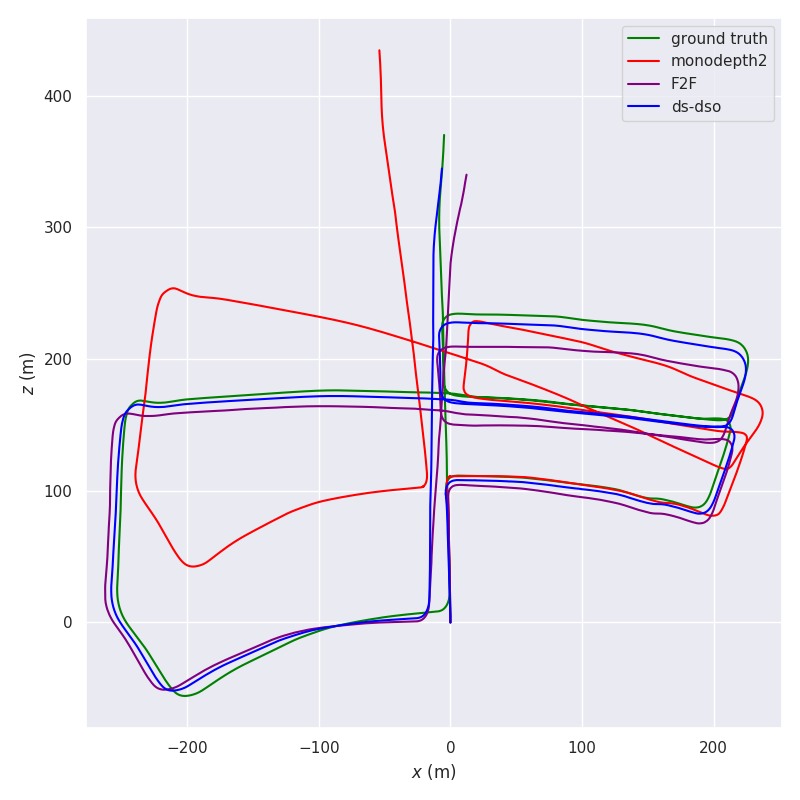} 
    \includegraphics[width=0.45\columnwidth]{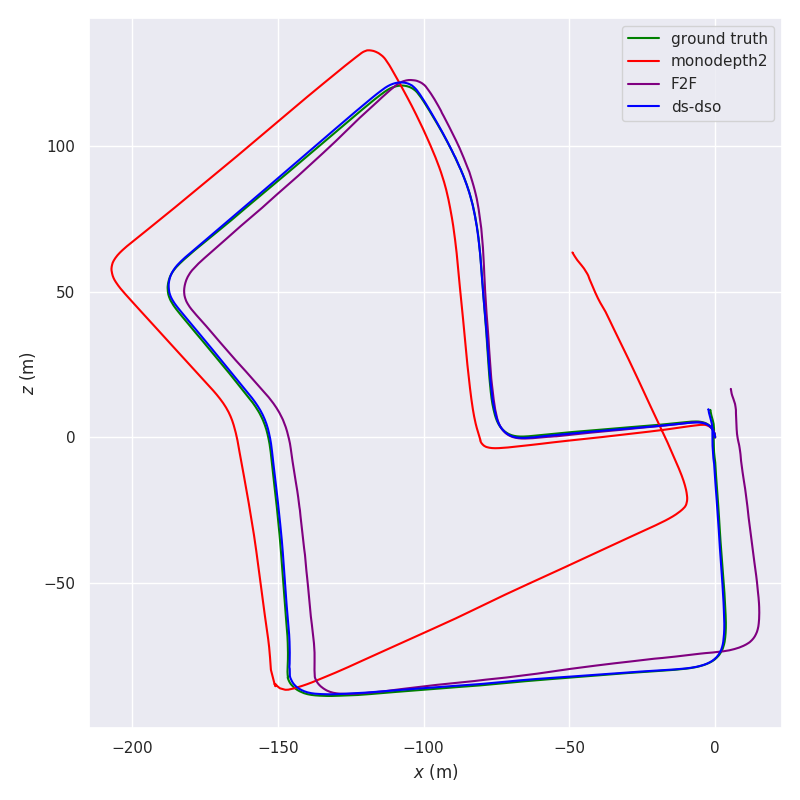}
}
\caption{\textbf{Qualitative trajectory estimation results on the KITTI Odometry Seq.~03,~04,~05 and 07}. We compare trajectory estimation results obtained via hand-engineered keypoint matching methods against our depth-aware learned keypoint matching, with a common visual odometry backend such as DSO. As illustrated in the figure, our self-supervised method is able to accurately and robustly track stable keypoints for the task of long-term trajectory estimation.} 
\label{fig:trajectory_evaluation}
\end{figure}

We show qualitative results of our method in Figure~\ref{fig:trajectory_evaluation}, noting that our DSO results accurately follow the ground truth trajectory with minimal scale drift.

\section{Keypoint Detector and Descriptor Evaluation Metrics}
\label{appendix:keypoint_metrics}

We follow~\cite{detone2018superpoint} and use the \textit{Repeatability} and \textit{Localization Error} metrics to estimate keypoint performance and \textit{Homography Accuracy} and \textit{Matching Score} matrics to estimate descriptor performance. We note that for all metrics we used a distance threshold of $3$. For the Homography estimation, consistent with other reported methods, we used $300$ keypoints with the highest scores. Similarly, for the frame to frame tracking we selected $480$ keypoints to estimate the relative pose.

\textbf{Repeatability} is computed as the ratio of correctly associated keypoints after warping onto the target frame. We consider a warped keypoint correctly associated if the nearest keypoint in the target frame (based on Euclidean distance) is below a certain threshold. 

\textbf{Localization Error} is computed as the average Euclidean distance between warped and associated keypoints. 

\textbf{Homography Accuracy} To compute the homography between two images we perform reciprocal descriptor matching and we used OpenCV’s \textit{findHomography} method with RANSAC, with a maximum of 5000 iterations and error threshold 3. To compute the Homography Accuracy we compare the estimated homography with the ground truth homography. Specifically we warp the image corners of the original image onto the target image using both the estimated homography and the ground truth homography, and we compute the average distance between the two sets of warped image corners, noting whether the average distance is below a certain threshold.  

\textbf{Matching Score} is computed as the ratio between successful keypoint associations between the two images, with the association being performed using Euclidean distance in descriptor space.

\section{Qualitative Results on HPatches}
\label{sec:qualitative_hpatches}

\begin{figure}[!h]
	\centering
	\includegraphics[width=0.9\textwidth]{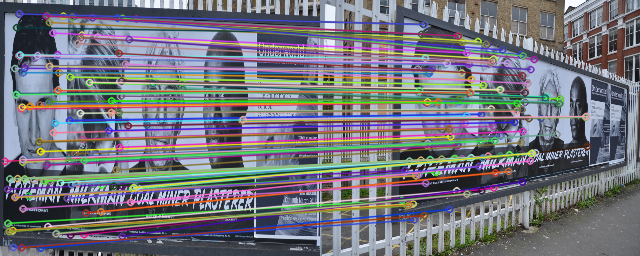}\\
	\includegraphics[width=0.9\textwidth]{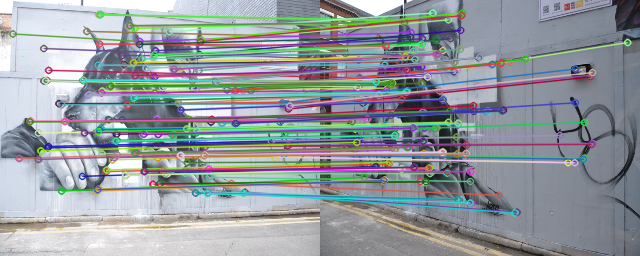}\\
	\includegraphics[width=0.9\textwidth]{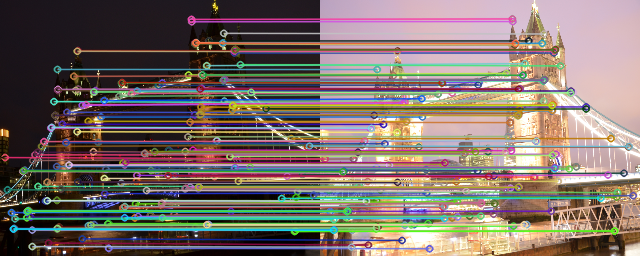}
	
	\caption{Qualitative matching results of our method on the HPatches dataset~\cite{balntas2017hpatches}.}
	\label{fig:qualitative_hpatches}
\end{figure}

Figure~\ref{fig:qualitative_hpatches} shows qualitative examples of our keypoints and matches on image pairs from the HPatches dataset~\cite{balntas2017hpatches}.

%% file: latex/results/depth_results.tex
\begin{table*}[t!]
\centering
{
\small
\begin{tabular}{c|lcccccccccc}
\toprule
& \textbf{Method} &
Abs Rel &
Sq Rel &
RMSE &
RMSE$_{log}$ &
$\delta < 1.25$ &
$\delta < 1.25^2$ &
$\delta < 1.25^3$\vspace{0.5mm}\\
\toprule
& Monodepth2~\cite{godard2018digging} & 
0.090 & 0.545 & 3.942 & 0.137 & 0.914 & 0.983 & 0.995\\
& DepthNet baseline & 
0.089 & 0.543 & 3.968 & 0.136 & 0.916 & 0.982 & 0.995 \\
& DepthNet finetuned & 
0.094 & 0.572 & 3.805 & 0.138 & 0.912 & 0.981 & 0.994 \\

\bottomrule

\end{tabular}
}
\caption{\textbf{Quantitative performance comparison of depth estimation on the KITTI dataset} for reported depths of up to 80m. For Abs Rel, Sq Rel, RMSE and RMSE$_{log}$ lower is better, and for $\delta < 1.25$, $\delta < 1.25^2$ and $\delta < 1.25^3$ higher is better.  All networks have been pre-trained on ImageNet~\cite{Deng09imagenet}. We evaluate on the annotated KITTI depth maps from \cite{uhrig2017sparsity}. At test-time, the scale for all the methods is corrected using the median ground-truth depth from the LiDAR.
}
\label{table:depth-accuracy}
\end{table*}

%% file: latex/073suppmat_dso.tex
\begin{figure}[!t]
\centering
\includegraphics[width=0.85\textwidth]{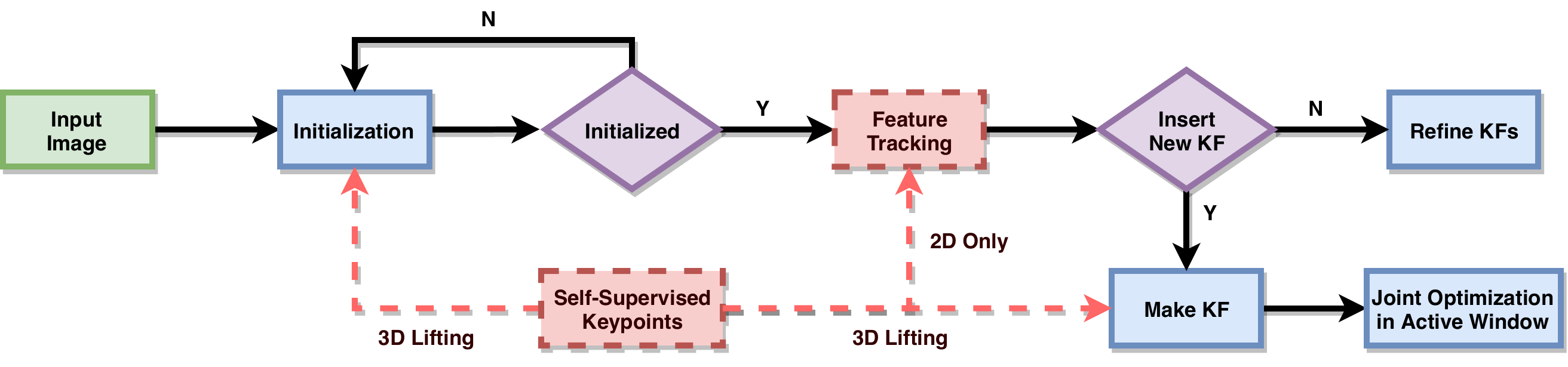}
\caption{\textbf{Direct Sparse Odometry (DSO) Integration}. We leverage our self-supervised depth-aware keypoint detection and description to improve the depth initialization and robust feature tracking components. The red block and arrows show that where the 3D keypoint is affecting the original DSO system, the purple texts show where 2D and 3D information is utilized.} 
\label{fig:ds_dso}
\end{figure}

\section{Direct Sparse Odometry (DSO) Integration}
\label{appendix:sec:ds-dso}

In this section, we explain how the fully self-supervised depth-aware keypoint network can be incorporated as the front-end into a visual SLAM framework.
We show that by integrating our method into a state-of-the-art monocular visual tracking framework such as DSO~\cite{engel2017direct}, we are able to achieve high accuracy long-term tracking results as reported in Table 1 of the main paper. 

Unlike other monocular visual odometry approaches, the superior keypoint matching and stable 3D lifting performance of our proposed method allows us to bootstrap the tracking system, rejecting  false  matches  and outliers and avoiding significant scale-drift.

Figure~\ref{fig:ds_dso} shows the whole pipeline of our Deep Semi-Direct Sparse Odometry (DS-DSO) system, which is built on top of the windowed sparse direct bundle adjustment formulation of DSO.
As illustrated, we improve depth-initialization of keyframes in the original DSO implmenetation by using the depth estimated through our proposed self-supervised 3D keypoint network. In addition, we modify the hand-engineered direct semi-dense tracking component
to our proposed sparse and robust learned keypoint-based method introduced in this work. 

%% file: latex/071suppmat_ssim.tex
We define the SSIM loss~\cite{wang2004image} as:

\begin{equation}
    SSIM(\mathbf{x},\mathbf{y}) = \frac{(2\mu_{x}\mu_{y} + C_1)(2\sigma_{xy} + C_2)}{(\mu_{x}^{2} + \mu_{y}^{2} + C_1)(\sigma_{x}^{2} + \sigma_{y}^{2}+C_2)}~,
  \label{eq:ssim}
\end{equation}

\noindent
with $C_1=1e^{-4}$ and $C_2=9e^{-4}$. To compute the per-patch mean and standard deviation $\mu_x$ and $\sigma_x$ we use a $3\times3$ block filter.

%% file: latex/072suppmat_pnp_pose.tex
Recall that we aim to minimize:

\begin{align} \label{eq:pnp_appendix}
E_{\psi}(X^{0}_{t \to c}) &= \left\lVert \bm{p}^{\phi}_c - \pi\left( X^0_{t \to c} \cdot \bm{P}^{\phi}_t  \right)   \right\rVert_2 \\
&= \left\lVert \bm{p}^{\phi}_c - \pi\left( R \cdot \bm{P}^{\phi}_t + t  \right)   \right\rVert_2~,
\end{align}

\noindent
where $R\in \mathbb{R}^{3 \times 3}$ is the rotation matrix and $t \in \mathbb{R}^3$ is the translation vector. They together compose a rigid body transform $\mathrm{exp}(\hat{x}) \in \mathbb{SE}(3)$, which is defined by $x = [\omega^T, t^T]^T\in \mathfrak{se}(3)$. $x$ is a member of the Lie algebra and is mapped to the Lie group $\mathbb{SE}(3)$ through the matrix exponential $\mathrm{exp}(.)$:
\begin{equation} \label{eq:warp}
\begin{gathered}
\mathrm{exp}({\hat{{x}}}) =  
\begin{bmatrix}
{R} &{t} \\
{0} & 1
\end{bmatrix} 
\quad,\quad
{\hat{{x}}} =
\begin{bmatrix}
[{\omega}]_\times & {t} \\
{0} & 1
\end{bmatrix} \\ 
\end{gathered}~,
\end{equation}

\noindent
where $[{\omega}]_\times$ is the skew-symmetric matrix of ${\omega}$.

The estimated relative pose can be obtained by optimizing the residual error in Equation~(\ref{eq:pnp_appendix}). The Gauss-Newton (GN) method is used to solve this non-linear least-squares problem. GN calculates ${x}$ iteratively as follows: 
\begin{equation}
{x}^{(n+1)} = {x}^{(n)} - \left (\mathbf{J_r}^{T} \mathbf{J_r} \right)^{-1} \mathbf{ J_r} ^\mathbf{T} \mathbf{r}({x}^{(n)})~,
\end{equation}
where $\mathbf{J_r}$ is the Jacobian matrix with respect to the residual measurements. RANSAC is performed to achieve a robust estimation and reject three major types of outliers which break the ego-motion assumption: false-positive matching pairs, dynamic objects or points with wrong depth estimations.

%% file: latex/results/appendix_traj_ablation_results.tex
\begin{table*}[!t]
\centering
\small
\renewcommand{\b}[1]{\textbf{#1}}
\renewcommand{\u}[1]{\underline{#1}}
\setlength{\tabcolsep}{.6em}

\resizebox{0.98\hsize}{!}{
\begin{tabular}{lcccccccccccccc}

\textbf{Method} & {Type} & 01$^*$ &    02$^*$  &  06$^*$  &  08$^*$   &   09$^*$  & 10$^*$   & 00$^\dagger$  & 03$^\dagger$  & 04$^\dagger$  & 05$^\dagger$  & 07$^\dagger$  & {Train} & {Test} \\
\toprule
& \multicolumn{12}{c}{$t_{rel}$ - Average Translational RMSE drift (\%) on trajectories of length 100-800m.}  \\
\midrule

Baseline
& Mono & 18.96 & 3.35 & 2.16 &  3.80 &  3.15 & 5.06      & 3.50 &  3.64 & 2.33 & 3.25 & 3.00 & 6.08 & 3.14 
\\
Ours $-$ Diff Pose
& Mono  & 20.17 & 3.37 & 2.15 &  3.01 &  2.61 & 5.39      & 2.89 &  3.10 & 2.88 & 3.09 & 2.66 & 6.12 & 2.92 
\\
Ours $-$ KPN trained
& Mono  & 19.09 & 3.30 & 2.23 &  3.16 &  2.84 & 5.03      & 2.83 &  3.29 & 2.02 & 3.69 & 2.58 & 5.94 & 2.88 
\\
Ours $-$ DN trained
& Mono  & 15.53 & 3.37 & 1.84 &  3.63 &  2.83 & 5.06      & 3.56 &  3.06 & 2.11 & 3.33 & 2.34 & 5.38 & 2.88 
\\
Ours 

& Mono  & 17.79 & 3.15 & 1.88 &  3.06 &  2.69 & 5.12      & 2.76 &  3.02 & 1.93 & 3.30 & 2.41 & 5.61 & 2.68 
\\
Ours + DSO
& Mono & 4.70 & 3.62 & 0.92 &  2.46 &  2.31 &  5.24 & 1.83 &  1.21 & 0.76 & 1.84 & 0.54 & 3.21 &  1.24
\\
\midrule
& \multicolumn{12}{c}{$r_{rel}$ - Average Rotational RMSE drift ($^{{\circ}}/100m$) on trajectories of length 100-800m.} \\ 
\midrule
Baseline
& Mono  & 1.02 & 1.12 & 0.82 &  1.00 &  0.72 & 1.43 &  1.26 &  3.17 & 1.09 & 1.24 & 1.39 & 1.02 & 1.63 
\\
Ours $-$ Diff Pose
& Mono  & 1.08 & 1.03 & 0.97 &  0.73 &  0.65 & 0.91 &  1.24 &  2.64 & 1.00 & 1.08 & 1.18 & 0.89 & 1.43 
\\
Ours $-$ KPN trained
& Mono  & 0.84 & 1.12 & 0.98 &  0.77 &  0.64 & 1.24 &  1.23 &  2.81 & 1.56 & 1.24 & 1.23 & 0.93 & 1.61 
\\
Ours $-$ DN trained
& Mono  & 0.66 & 1.13 & 0.68 &  0.88 &  0.62 & 1.44 &  1.20 &  2.74 & 1.73 & 1.22 & 1.04 & 0.91 & 1.58 
\\
Ours
& Mono  & 0.72 & 1.01 & 0.80 &  0.76 &  0.61 & 1.07 &  1.17 &  2.45 & 1.93 & 1.11 & 1.16 & 0.83 & 1.56 
\\
Ours + DSO
& Mono   & 0.16 & 0.22 & 0.13 & 0.31 &  0.30 & 0.29 &  0.33 & 0.33 & 0.18  & 0.22 &  0.23 & 0.24 &  0.26
\\
\bottomrule
\end{tabular}}
\vspace{1mm}
\caption{\textbf{Detailed results of our Pose Ablative Analysis.} Note: our results are obtained after performing a single Sim(3) alignment step~\cite{grupp2017evo} wrt. the ground truth trajectories. 
}
\label{table:kitti-eigen-traj-appendix} 
\end{table*}

%% file: latex/results/supp_kitti_odom_results.tex
\begin{table}[!h]
\centering
\small
\renewcommand{\b}[1]{\textbf{#1}}
\renewcommand{\u}[1]{\underline{#1}}
\setlength{\tabcolsep}{.6em}

\begin{tabular}{lccc}

&  SfMLearner~\cite{zhou2017unsupervised} & Zhan et al~\cite{zhan2018unsupervised}  & Ours \\

\toprule

& \multicolumn{3}{c}{$t_{rel}$ - Average Translational RMSE drift (\%).}  \\
\midrule

Seq 09 & 18.8 & 11.9 & 3.14 \\
Seq 10 & 14.3 & 12.6 & 5.45 \\
Mean   & 16.6 & 12.3 & 4.30 \\

\midrule
& \multicolumn{3}{c}{$r_{rel}$ - Average Rotational RMSE drift ($^{{\circ}}/100m$).} \\ 
\midrule

Seq 09 & 3.21 & 3.60 & 0.73 \\
Seq 10 & 3.30 & 3.43 & 1.18 \\
Mean   & 3.26 & 3.52 & 0.90  \\

\bottomrule
\end{tabular}
\vspace{1mm}
\caption{\textbf{Comparison of vision-based trajectory estimation with state-of-the-art methods.}~The \textit{Type} column indicates the data used at training time. Note: All methods are evaluated on monocular data. Our results are obtained after performing a single Sim(3) alignment step~\cite{grupp2017evo} wrt. the ground truth trajectories. Training seq. 00-08; test seq: 09 and 10. 
}
\label{table:appendix-kitti-odom-results-horizontal} 
\end{table}

%% file: latex/results/supp_kitti_odom_ours_detailed.tex
\begin{table*}[!h]
\centering
\small
\renewcommand{\b}[1]{\textbf{#1}}
\renewcommand{\u}[1]{\underline{#1}}
\setlength{\tabcolsep}{.6em}

\resizebox{0.98\hsize}{!}{
\begin{tabular}{lccccccccccccc}

\textbf{Metric/Seq}  & 00 &  01  &  02  &  03  &   04  & 05  & 06  & 07  & 08  & 09  & 10  & {Train} & {Test} \\
\toprule
$t_{rel}$ & 4.11 &	47.21 &	3.85 &	3.62 &	2.90 &	3.04	& 1.84 & 	2.21	& 3.82 &	3.14 &	5.45 &	8.07 &	4.30
\\

\midrule

$r_{rel}$ & 1.38 &	1.51 &	1.09 &	3.53 &	1.08 &	0.94 &	0.68 &	1.09 &	1.00 &	0.73 &	1.18 &	1.37 &	0.9
\\

\bottomrule
\end{tabular}}
\vspace{1mm}
\caption{\textbf{Detailed results of our method on the KITTI odometry sequences.}~Training seq. 00-08; test seq: 09 and 10. ~Our results are obtained after performing a single Sim(3) alignment step~\cite{grupp2017evo} wrt. the ground truth trajectories.}
\label{table:supp-kitti-odom-results-ours-detailed} 
\end{table*}

%% file: main.bbl
\begin{thebibliography}{58}
\providecommand{\natexlab}[1]{#1}
\providecommand{\url}[1]{\texttt{#1}}
\expandafter\ifx\csname urlstyle\endcsname\relax
  \providecommand{\doi}[1]{doi: #1}\else
  \providecommand{\doi}{doi: \begingroup \urlstyle{rm}\Url}\fi

\bibitem[Agarwal et~al.(2010)Agarwal, Snavely, Seitz, and
  Szeliski]{agarwal2010bundle}
S.~Agarwal, N.~Snavely, S.~M. Seitz, and R.~Szeliski.
\newblock Bundle adjustment in the large.
\newblock In \emph{European conference on computer vision}, pages 29--42.
  Springer, 2010.

\bibitem[Cadena et~al.(2016)Cadena, Carlone, Carrillo, Latif, Scaramuzza,
  Neira, Reid, and Leonard]{cadena2016past}
C.~Cadena, L.~Carlone, H.~Carrillo, Y.~Latif, D.~Scaramuzza, J.~Neira, I.~Reid,
  and J.~J. Leonard.
\newblock Past, present, and future of simultaneous localization and mapping:
  Toward the robust-perception age.
\newblock \emph{IEEE Transactions on robotics}, 32\penalty0 (6):\penalty0
  1309--1332, 2016.

\bibitem[Lowe(1999)]{lowe1999object}
D.~G. Lowe.
\newblock Object recognition from local scale-invariant features.
\newblock In \emph{Proceedings of the seventh IEEE international conference on
  computer vision}, volume~2, pages 1150--1157. Ieee, 1999.

\bibitem[Rublee et~al.(2011)Rublee, Rabaud, Konolige, and
  Bradski]{rublee2011orb}
E.~Rublee, V.~Rabaud, K.~Konolige, and G.~Bradski.
\newblock Orb: An efficient alternative to sift or surf.
\newblock In \emph{2011 International conference on computer vision}, pages
  2564--2571. Ieee, 2011.

\bibitem[Sarlin et~al.(2019)Sarlin, Cadena, Siegwart, and
  Dymczyk]{sarlin2019coarse}
P.-E. Sarlin, C.~Cadena, R.~Siegwart, and M.~Dymczyk.
\newblock From coarse to fine: Robust hierarchical localization at large scale.
\newblock In \emph{Proceedings of the IEEE Conference on Computer Vision and
  Pattern Recognition}, pages 12716--12725, 2019.

\bibitem[He et~al.(2016)He, Zhang, Ren, and Sun]{he2016deep}
K.~He, X.~Zhang, S.~Ren, and J.~Sun.
\newblock Deep residual learning for image recognition.
\newblock In \emph{Proceedings of the IEEE conference on computer vision and
  pattern recognition}, pages 770--778, 2016.

\bibitem[Alp~G{\"u}ler et~al.(2018)Alp~G{\"u}ler, Neverova, and
  Kokkinos]{alp2018densepose}
R.~Alp~G{\"u}ler, N.~Neverova, and I.~Kokkinos.
\newblock Densepose: Dense human pose estimation in the wild.
\newblock In \emph{Proceedings of the IEEE Conference on Computer Vision and
  Pattern Recognition}, pages 7297--7306, 2018.

\bibitem[Kirillov et~al.(2019)Kirillov, Girshick, He, and
  Doll{\'a}r]{kirillov2019panoptic}
A.~Kirillov, R.~Girshick, K.~He, and P.~Doll{\'a}r.
\newblock Panoptic feature pyramid networks.
\newblock In \emph{Proceedings of the IEEE Conference on Computer Vision and
  Pattern Recognition}, pages 6399--6408, 2019.

\bibitem[DeTone et~al.(2018)DeTone, Malisiewicz, and
  Rabinovich]{detone2018superpoint}
D.~DeTone, T.~Malisiewicz, and A.~Rabinovich.
\newblock Superpoint: Self-supervised interest point detection and description.
\newblock In \emph{Proceedings of the IEEE Conference on Computer Vision and
  Pattern Recognition Workshops}, pages 224--236, 2018.

\bibitem[Christiansen et~al.(2019)Christiansen, Kragh, Brodskiy, and
  Karstoft]{christiansen2019unsuperpoint}
P.~H. Christiansen, M.~F. Kragh, Y.~Brodskiy, and H.~Karstoft.
\newblock Unsuperpoint: End-to-end unsupervised interest point detector and
  descriptor.
\newblock \emph{arXiv preprint arXiv:1907.04011}, 2019.

\bibitem[Revaud et~al.(2019)Revaud, Weinzaepfel, De~Souza, Pion, Csurka, Cabon,
  and Humenberger]{revaud2019r2d2}
J.~Revaud, P.~Weinzaepfel, C.~De~Souza, N.~Pion, G.~Csurka, Y.~Cabon, and
  M.~Humenberger.
\newblock R2d2: Repeatable and reliable detector and descriptor.
\newblock \emph{arXiv preprint arXiv:1906.06195}, 2019.

\bibitem[Bhowmik et~al.(2020)Bhowmik, Gumhold, Rother, and
  Brachmann]{bhowmik2020reinforced}
A.~Bhowmik, S.~Gumhold, C.~Rother, and E.~Brachmann.
\newblock Reinforced feature points: Optimizing feature detection and
  description for a high-level task.
\newblock In \emph{Proceedings of the IEEE/CVF Conference on Computer Vision
  and Pattern Recognition}, pages 4948--4957, 2020.

\bibitem[Engel et~al.(2017)Engel, Koltun, and Cremers]{engel2017direct}
J.~Engel, V.~Koltun, and D.~Cremers.
\newblock Direct sparse odometry.
\newblock \emph{IEEE transactions on pattern analysis and machine
  intelligence}, 40\penalty0 (3):\penalty0 611--625, 2017.

\bibitem[Yi et~al.(2016)Yi, Trulls, Lepetit, and Fua]{yi2016lift}
K.~M. Yi, E.~Trulls, V.~Lepetit, and P.~Fua.
\newblock Lift: Learned invariant feature transform.
\newblock In \emph{European Conference on Computer Vision}, pages 467--483.
  Springer, 2016.

\bibitem[Ono et~al.(2018)Ono, Trulls, Fua, and Yi]{ono2018lf}
Y.~Ono, E.~Trulls, P.~Fua, and K.~M. Yi.
\newblock Lf-net: learning local features from images.
\newblock In \emph{Advances in neural information processing systems}, pages
  6234--6244, 2018.

\bibitem[Suwajanakorn et~al.(2018)Suwajanakorn, Snavely, Tompson, and
  Norouzi]{suwajanakorn2018discovery}
S.~Suwajanakorn, N.~Snavely, J.~J. Tompson, and M.~Norouzi.
\newblock Discovery of latent 3d keypoints via end-to-end geometric reasoning.
\newblock In \emph{Advances in neural information processing systems}, pages
  2059--2070, 2018.

\bibitem[Tang et~al.(2019)Tang, Kim, Guizilini, Pillai, and
  Ambrus]{tang2019neural}
J.~Tang, H.~Kim, V.~Guizilini, S.~Pillai, and R.~Ambrus.
\newblock Neural outlier rejection for self-supervised keypoint learning.
\newblock In \emph{International Conference on Learning Representations}, 2019.

\bibitem[Sarlin et~al.(2019)Sarlin, Cadena, Siegwart, and
  Dymczyk]{Sarlin:etal:CVPR2019}
P.-E. Sarlin, C.~Cadena, R.~Siegwart, and M.~Dymczyk.
\newblock From coarse to fine: Robust hierarchical localization at large scale.
\newblock In \emph{The IEEE Conference on Computer Vision and Pattern
  Recognition (CVPR)}, June 2019.

\bibitem[Sarlin et~al.(2020)Sarlin, DeTone, Malisiewicz, and
  Rabinovich]{sarlin2020superglue}
P.-E. Sarlin, D.~DeTone, T.~Malisiewicz, and A.~Rabinovich.
\newblock Superglue: Learning feature matching with graph neural networks.
\newblock In \emph{Proceedings of the IEEE/CVF Conference on Computer Vision
  and Pattern Recognition}, pages 4938--4947, 2020.

\bibitem[Bai et~al.(2020)Bai, Luo, Zhou, Fu, Quan, and Tai]{bai2020d3feat}
X.~Bai, Z.~Luo, L.~Zhou, H.~Fu, L.~Quan, and C.-L. Tai.
\newblock D3feat: Joint learning of dense detection and description of 3d local
  features.
\newblock In \emph{Proceedings of the IEEE/CVF Conference on Computer Vision
  and Pattern Recognition}, pages 6359--6367, 2020.

\bibitem[Godard et~al.(2017)Godard, Mac~Aodha, and
  Brostow]{godard2017unsupervised}
C.~Godard, O.~Mac~Aodha, and G.~J. Brostow.
\newblock Unsupervised monocular depth estimation with left-right consistency.
\newblock In \emph{Proceedings of the IEEE Conference on Computer Vision and
  Pattern Recognition}, pages 270--279, 2017.

\bibitem[Zhou et~al.(2017)Zhou, Brown, Snavely, and Lowe]{zhou2017unsupervised}
T.~Zhou, M.~Brown, N.~Snavely, and D.~G. Lowe.
\newblock Unsupervised learning of depth and ego-motion from video.
\newblock In \emph{Proceedings of the IEEE Conference on Computer Vision and
  Pattern Recognition}, pages 1851--1858, 2017.

\bibitem[Pillai et~al.(2019)Pillai, Ambru{\c{s}}, and
  Gaidon]{pillai2018superdepth}
S.~Pillai, R.~Ambru{\c{s}}, and A.~Gaidon.
\newblock Superdepth: Self-supervised, super-resolved monocular depth
  estimation.
\newblock In \emph{2019 International Conference on Robotics and Automation
  (ICRA)}, pages 9250--9256. IEEE, 2019.

\bibitem[Casser et~al.(2019)Casser, Pirk, Mahjourian, and
  Angelova]{casser2018depth}
V.~Casser, S.~Pirk, R.~Mahjourian, and A.~Angelova.
\newblock Depth prediction without the sensors: Leveraging structure for
  unsupervised learning from monocular videos.
\newblock In \emph{Proceedings of the AAAI Conference on Artificial
  Intelligence}, volume~33, pages 8001--8008, 2019.

\bibitem[Bian et~al.(2019)Bian, Li, Wang, Zhan, Shen, Cheng, and
  Reid]{bian2019unsupervised}
J.~Bian, Z.~Li, N.~Wang, H.~Zhan, C.~Shen, M.-M. Cheng, and I.~Reid.
\newblock Unsupervised scale-consistent depth and ego-motion learning from
  monocular video.
\newblock In \emph{Advances in neural information processing systems}, pages
  35--45, 2019.

\bibitem[Ambrus et~al.(2019)Ambrus, Guizilini, Li, Pillai, and
  Gaidon]{ambrus2019two}
R.~Ambrus, V.~Guizilini, J.~Li, S.~Pillai, and A.~Gaidon.
\newblock Two stream networks for self-supervised ego-motion estimation.
\newblock In \emph{Proceedings of the 3rd International Conference on Robot
  Learning (CoRL)}, 2019.

\bibitem[Luo et~al.(2019)Luo, Yang, Wang, Wang, Xu, Nevatia, and
  Yuille]{luo2019every}
C.~Luo, Z.~Yang, P.~Wang, Y.~Wang, W.~Xu, R.~Nevatia, and A.~Yuille.
\newblock Every pixel counts++: Joint learning of geometry and motion with 3d
  holistic understanding.
\newblock \emph{IEEE Transactions on Pattern Analysis and Machine
  Intelligence}, 2019.

\bibitem[Zhan et~al.(2020)Zhan, Weerasekera, Bian, and Reid]{zhan2019visual}
H.~Zhan, C.~S. Weerasekera, J.-W. Bian, and I.~Reid.
\newblock Visual odometry revisited: What should be learnt?
\newblock In \emph{2020 IEEE International Conference on Robotics and
  Automation (ICRA)}, pages 4203--4210. IEEE, 2020.

\bibitem[Zhao et~al.(2020)Zhao, Liu, Shu, and Liu]{zhao2020towards}
W.~Zhao, S.~Liu, Y.~Shu, and Y.-J. Liu.
\newblock Towards better generalization: Joint depth-pose learning without
  posenet.
\newblock In \emph{Proceedings of the IEEE/CVF Conference on Computer Vision
  and Pattern Recognition}, pages 9151--9161, 2020.

\bibitem[Gordon et~al.(2019)Gordon, Li, Jonschkowski, and
  Angelova]{gordon2019depth}
A.~Gordon, H.~Li, R.~Jonschkowski, and A.~Angelova.
\newblock Depth from videos in the wild: Unsupervised monocular depth learning
  from unknown cameras.
\newblock In \emph{Proceedings of the IEEE International Conference on Computer
  Vision}, pages 8977--8986, 2019.

\bibitem[Feng and Gu(2019)]{feng2019sganvo}
T.~Feng and D.~Gu.
\newblock Sganvo: Unsupervised deep visual odometry and depth estimation with
  stacked generative adversarial networks.
\newblock \emph{IEEE Robotics and Automation Letters}, 4\penalty0 (4):\penalty0
  4431--4437, 2019.

\bibitem[Yang et~al.(2018)Yang, Wang, Stuckler, and Cremers]{yang2018deep}
N.~Yang, R.~Wang, J.~Stuckler, and D.~Cremers.
\newblock Deep virtual stereo odometry: Leveraging deep depth prediction for
  monocular direct sparse odometry.
\newblock In \emph{Proceedings of the European Conference on Computer Vision
  (ECCV)}, pages 817--833, 2018.

\bibitem[Yang et~al.(2020)Yang, Stumberg, Wang, and Cremers]{yang2020d3vo}
N.~Yang, L.~v. Stumberg, R.~Wang, and D.~Cremers.
\newblock D3vo: Deep depth, deep pose and deep uncertainty for monocular visual
  odometry.
\newblock In \emph{Proceedings of the IEEE/CVF Conference on Computer Vision
  and Pattern Recognition}, pages 1281--1292, 2020.

\bibitem[Wang and Solomon(2019)]{wang2019deep}
Y.~Wang and J.~M. Solomon.
\newblock Deep closest point: Learning representations for point cloud
  registration.
\newblock In \emph{Proceedings of the IEEE International Conference on Computer
  Vision}, pages 3523--3532, 2019.

\bibitem[Choy et~al.(2020)Choy, Dong, and Koltun]{choy2020deep}
C.~Choy, W.~Dong, and V.~Koltun.
\newblock Deep global registration.
\newblock In \emph{Proceedings of the IEEE/CVF Conference on Computer Vision
  and Pattern Recognition}, pages 2514--2523, 2020.

\bibitem[Krishna~Murthy et~al.(2020)Krishna~Murthy, Iyer, and
  Paull]{murthy2019gradslam}
J.~Krishna~Murthy, G.~Iyer, and L.~Paull.
\newblock gradslam: Dense slam meets automatic differentiation.
\newblock \emph{ICRA (to appear)}, 2020.

\bibitem[Lepetit et~al.(2009)Lepetit, Moreno-Noguer, and Fua]{lepetit2009epnp}
V.~Lepetit, F.~Moreno-Noguer, and P.~Fua.
\newblock Epnp: An accurate o (n) solution to the pnp problem.
\newblock \emph{International journal of computer vision}, 81\penalty0
  (2):\penalty0 155, 2009.

\bibitem[Sheffer and Wiesel(2020)]{sheffer2020pnp}
R.~Sheffer and A.~Wiesel.
\newblock Pnp-net: A hybrid perspective-n-point network.
\newblock \emph{arXiv preprint arXiv:2003.04626}, 2020.

\bibitem[Zhang(2000)]{zhang2000flexible}
Z.~Zhang.
\newblock A flexible new technique for camera calibration.
\newblock \emph{IEEE Transactions on pattern analysis and machine
  intelligence}, 22\penalty0 (11):\penalty0 1330--1334, 2000.

\bibitem[Godard et~al.(2019)Godard, Mac~Aodha, Firman, and
  Brostow]{godard2018digging}
C.~Godard, O.~Mac~Aodha, M.~Firman, and G.~J. Brostow.
\newblock Digging into self-supervised monocular depth estimation.
\newblock In \emph{Proceedings of the IEEE international conference on computer
  vision}, pages 3828--3838, 2019.

\bibitem[Guizilini et~al.(2020)Guizilini, Ambrus, Pillai, Raventos, and
  Gaidon]{guizilini2019packnet}
V.~Guizilini, R.~Ambrus, S.~Pillai, A.~Raventos, and A.~Gaidon.
\newblock 3d packing for self-supervised monocular depth estimation.
\newblock In \emph{Proceedings of the IEEE/CVF Conference on Computer Vision
  and Pattern Recognition}, pages 2485--2494, 2020.

\bibitem[Jaderberg et~al.(2015)Jaderberg, Simonyan, Zisserman,
  et~al.]{jaderberg2015spatial}
M.~Jaderberg, K.~Simonyan, A.~Zisserman, et~al.
\newblock Spatial transformer networks.
\newblock In \emph{Advances in neural information processing systems}, pages
  2017--2025, 2015.

\bibitem[Wang et~al.(2004)Wang, Bovik, Sheikh, and Simoncelli]{wang2004image}
Z.~Wang, A.~C. Bovik, H.~R. Sheikh, and E.~P. Simoncelli.
\newblock Image quality assessment: from error visibility to structural
  similarity.
\newblock \emph{IEEE transactions on image processing}, 13\penalty0
  (4):\penalty0 600--612, 2004.

\bibitem[Geiger et~al.(2013)Geiger, Lenz, Stiller, and
  Urtasun]{geiger2013vision}
A.~Geiger, P.~Lenz, C.~Stiller, and R.~Urtasun.
\newblock {Vision meets robotics: The KITTI dataset}.
\newblock \emph{The International Journal of Robotics Research}, 32\penalty0
  (11):\penalty0 1231--1237, 2013.

\bibitem[Grupp(2017)]{grupp2017evo}
M.~Grupp.
\newblock evo: Python package for the evaluation of odometry and slam.
\newblock \url{https://github.com/MichaelGrupp/evo}, 2017.

\bibitem[Eigen et~al.(2014)Eigen, Puhrsch, and Fergus]{eigen2014depth}
D.~Eigen, C.~Puhrsch, and R.~Fergus.
\newblock Depth map prediction from a single image using a multi-scale deep
  network.
\newblock In \emph{Advances in neural information processing systems}, pages
  2366--2374, 2014.

\bibitem[Balntas et~al.(2017)Balntas, Lenc, Vedaldi, and
  Mikolajczyk]{balntas2017hpatches}
V.~Balntas, K.~Lenc, A.~Vedaldi, and K.~Mikolajczyk.
\newblock Hpatches: A benchmark and evaluation of handcrafted and learned local
  descriptors.
\newblock In \emph{Proceedings of the IEEE Conference on Computer Vision and
  Pattern Recognition}, pages 5173--5182, 2017.

\bibitem[Lin et~al.(2014)Lin, Maire, Belongie, Hays, Perona, Ramanan,
  Doll{\'a}r, and Zitnick]{lin2014microsoft}
T.-Y. Lin, M.~Maire, S.~Belongie, J.~Hays, P.~Perona, D.~Ramanan,
  P.~Doll{\'a}r, and C.~L. Zitnick.
\newblock Microsoft coco: Common objects in context.
\newblock In \emph{European conference on computer vision}, pages 740--755.
  Springer, 2014.

\bibitem[Paszke et~al.(2017)Paszke, Gross, Chintala, Chanan, Yang, DeVito, Lin,
  Desmaison, Antiga, and Lerer]{paszke2017automatic}
A.~Paszke, S.~Gross, S.~Chintala, G.~Chanan, E.~Yang, Z.~DeVito, Z.~Lin,
  A.~Desmaison, L.~Antiga, and A.~Lerer.
\newblock Automatic differentiation in pytorch.
\newblock In \emph{NIPS-W}, 2017.

\bibitem[Kingma and Ba(2014)]{kingma2014adam}
D.~P. Kingma and J.~Ba.
\newblock Adam: A method for stochastic optimization.
\newblock \emph{arXiv preprint arXiv:1412.6980}, 2014.

\bibitem[Deng et~al.(2009)Deng, Dong, Socher, jia Li, Li, and
  Fei-fei]{Deng09imagenet}
J.~Deng, W.~Dong, R.~Socher, L.~jia Li, K.~Li, and L.~Fei-fei.
\newblock Imagenet: A large-scale hierarchical image database.
\newblock In \emph{Proceedings of the IEEE Conference on Computer Vision and
  Pattern Recognition}, 2009.

\bibitem[Mur-Artal and Tard{\'o}s(2017)]{mur2017orb}
R.~Mur-Artal and J.~D. Tard{\'o}s.
\newblock Orb-slam2: An open-source slam system for monocular, stereo, and
  rgb-d cameras.
\newblock \emph{IEEE Transactions on Robotics}, 33\penalty0 (5):\penalty0
  1255--1262, 2017.

\bibitem[Engel et~al.(2014)Engel, Sch{\"o}ps, and Cremers]{engel2014lsd}
J.~Engel, T.~Sch{\"o}ps, and D.~Cremers.
\newblock Lsd-slam: Large-scale direct monocular slam.
\newblock In \emph{European conference on computer vision}, pages 834--849.
  Springer, 2014.

\bibitem[Li et~al.(2018)Li, Wang, Long, and Gu]{li2017undeepvo}
R.~Li, S.~Wang, Z.~Long, and D.~Gu.
\newblock Undeepvo: Monocular visual odometry through unsupervised deep
  learning.
\newblock In \emph{2018 IEEE international conference on robotics and
  automation (ICRA)}, pages 7286--7291. IEEE, 2018.

\bibitem[Bay et~al.(2006)Bay, Tuytelaars, and Van~Gool]{bay2006surf}
H.~Bay, T.~Tuytelaars, and L.~Van~Gool.
\newblock Surf: Speeded up robust features.
\newblock In \emph{European conference on computer vision}, pages 404--417.
  Springer, 2006.

\bibitem[Leutenegger et~al.(2011)Leutenegger, Chli, and
  Siegwart]{leutenegger2011brisk}
S.~Leutenegger, M.~Chli, and R.~Siegwart.
\newblock Brisk: Binary robust invariant scalable keypoints.
\newblock In \emph{2011 IEEE international conference on computer vision
  (ICCV)}, pages 2548--2555. Ieee, 2011.

\bibitem[Uhrig et~al.(2017)Uhrig, Schneider, Schneider, Franke, Brox, and
  Geiger]{uhrig2017sparsity}
J.~Uhrig, N.~Schneider, L.~Schneider, U.~Franke, T.~Brox, and A.~Geiger.
\newblock Sparsity invariant cnns.
\newblock In \emph{2017 International Conference on 3D Vision (3DV)}, pages
  11--20. IEEE, 2017.

\bibitem[Zhan et~al.(2018)Zhan, Garg, Saroj~Weerasekera, Li, Agarwal, and
  Reid]{zhan2018unsupervised}
H.~Zhan, R.~Garg, C.~Saroj~Weerasekera, K.~Li, H.~Agarwal, and I.~Reid.
\newblock Unsupervised learning of monocular depth estimation and visual
  odometry with deep feature reconstruction.
\newblock In \emph{Proceedings of the IEEE Conference on Computer Vision and
  Pattern Recognition}, pages 340--349, 2018.

\end{thebibliography}
